%% file: main.tex
\documentclass[10pt,twocolumn,letterpaper]{article}

\usepackage{cvpr}              


\definecolor{cvprblue}{rgb}{0.21,0.49,0.74}
\usepackage[pagebackref,breaklinks,colorlinks,allcolors=cvprblue]{hyperref}
\usepackage{booktabs}
\usepackage{enumitem}
\usepackage{array,multirow}
\usepackage{float}
\usepackage{arydshln}
\usepackage{wasysym}

\def\paperID{11} 
\def\confName{CVPR}
\def\confYear{2025}

\title{Leveraging Anthropometric Measurements to Improve Human Mesh Estimation and Ensure Consistent Body Shapes}

\author{Katja Ludwig, Julian Lorenz, Daniel Kienzle, Tuan Bui \& Rainer Lienhart\\
Chair for Machine Learning \& Computer Vision, University of Augsburg, Germany\\
{\tt\small\{firstname.lastname\}@uni-a.de} \\
}

\begin{document}
\maketitle
\input{sec/0_abstract}    
\input{sec/1_intro}
\input{sec/2_related_work}
\input{sec/3_3d_gt_errors}
\input{sec/4_a2b}
\input{sec/5_a2b_hme}
\input{sec/6_conclusion}
{
    \small
    \bibliographystyle{ieeenat_fullname}
    \bibliography{main}
}

\input{sec/X_suppl}

\end{document}

%% file: sec/0_abstract.tex
\begin{abstract}
The basic body shape (i.e., the body shape in T-pose) of a person does not change within a single video. However, most SOTA human mesh estimation (HME) models output a slightly different, thus inconsistent basic body shape for each video frame. Furthermore, we find that SOTA 3D human pose estimation (HPE) models outperform HME models regarding the precision of the estimated 3D keypoint positions. We solve the problem of inconsistent body shapes by leveraging anthropometric measurements like taken by tailors from humans. We create a model called A2B that converts given anthropometric measurements to basic body shape parameters of human mesh models. We obtain superior and consistent human meshes by combining the A2B model results with the keypoints of 3D HPE models using inverse kinematics. We evaluate our approach on challenging datasets like ASPset or fit3D, where we can lower the MPJPE by over 30\,mm compared to SOTA HME models. Further, replacing estimates of the body shape parameters from existing HME models with A2B results not only increases the performance of these HME models, but also guarantees consistent body shapes.
\end{abstract}

%% file: sec/1_intro.tex
\section{Introduction}
\label{sec:intro}

Creating an accurate 3D human mesh from monocular images or videos creates new opportunities in fields like 3D animation, gaming, fashion, sports, etc. In many of these application fields, videos are of main interest. 
\begin{figure}[htb]
  \centering
    \includegraphics[width=0.8\linewidth]{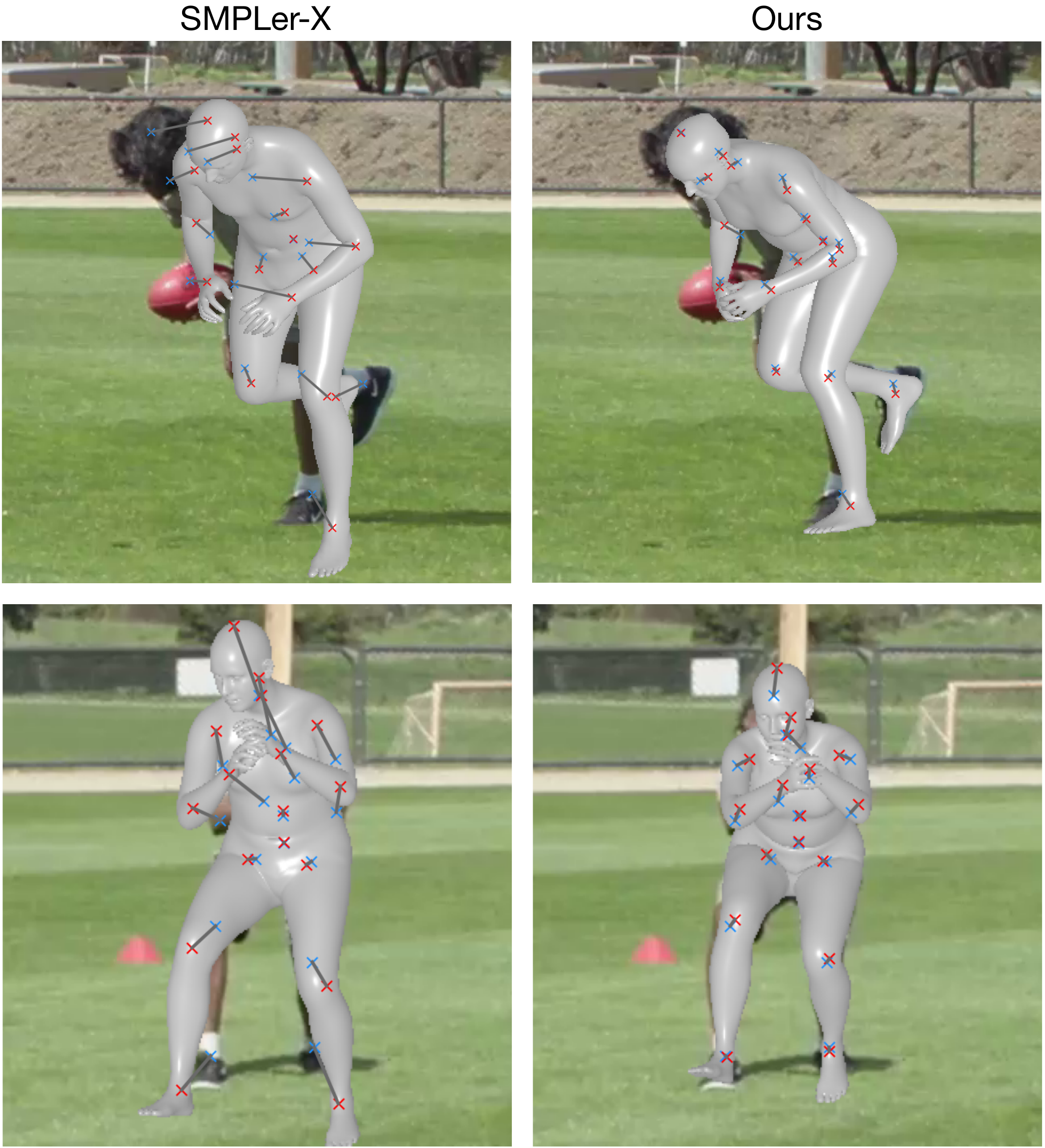}
   \caption{Two qualitative examples from the ASPset sports dataset. The result from a SOTA HME model, SMPLer-X \citep{smplerx}, is shown on the left, the result from our model on the right, respectively. \textcolor{blue}{GT joints} and \textcolor{red}{estimated joints} are color-coded. Corresponding joints are connected.}
   \label{fig:vis}
   \vspace{-0.2cm}
\end{figure}
While applying HME to videos, 
analyses of results of SOTA HME models show that the basic body shape of the meshes of the same person differs from frame to frame. \footnote{The body shape in a given pose is usually modeled by a basic body shape (given in T-pose) plus an additional pose-dependent deformation. We call the basic body shape just body shape in the following, as the pose-specific correction is computed from the pose and does not need to be estimated.} 
Worse, an analysis of currently used 3D mesh and pose datasets reveals the same inconsistencies in the provided ground truth (GT) data. For a precise body posture analysis, as it is necessary in many sports disciplines, an exact model of the athlete's body shape is required. Therefore, most professional athletes are measured anthropometrically during performance assessments today. Moreover, the body shape of an actor performing motions for 3D animations needs to be consistent as the basic body shapes does not change during performances. Thus, the changing body shapes of HME models for the same person are highly unwanted and simply wrong. 

Our work aims to create a single perfectly fitting basic body shape for each human and reuse it for all videos with this person. Measuring the human body has already been done for centuries to fit suits or dresses perfectly to a specific body shape. In many applications, measuring the person in action beforehand would add only a marginal overhead, but improves the results dramatically. For this reason we propose to use these measurements. Body shape parameters of common human mesh models like SMPL-X \citep{smplx} are not human interpretable. Therefore, it is not possible to obtain the perfect body shape parameters by anthropometric measurements. Hence, we train a machine learning model (called \emph{A2B}, \underline{a}nthropometric measurements \underline{to} \underline{b}ody shape) to translate those measurements into body shape parameters for HME. With this approach, measuring a person once creates the body shape that can be used for all frames in all evaluation videos. 

HME models are performing well on everyday data. However, in more challenging scenarios like sports, their performance is inferior to fine-tuned SOTA 3D HPE models. 3D HPE models only predict 3D keypoints resulting in a stick-figure pose, whereas HME models output a posed mesh including the human's surface. Due to the lack of GT meshes, HME models cannot be trained on datasets with solely 3D keypoint annotations. The usage of synthetic data is emerging in the field, but is not applicable to challenging or specific scenarios like sports. In this paper, we propose a solution to that problem. With our A2B model and anthropometric measurements, we can now create the body shape parameters of humans needed for HME. We further apply inverse kinematics (IK) to produce the rotations that are missing in the 3D stick-figure model that is created by 3D HPE models. Together with our A2B body shape, we are able to generate human meshes that have a consistent body shape and a precise pose. The main goal of this paper is to \textbf{estimate the best possible human mesh with a consistent body shape by adding the marginal overhead of measuring humans}. Since our main focus is sports, this overhead is negligible, as professional athletes are commonly measured anyway. We show qualitative examples of our model and a SOTA HME model, SMPLer-X \citep{smplerx}, in Figure \ref{fig:vis}.
Our approach is generally applicable to any HME problem. We choose sports datasets to validate our proposed approach, since the poses in sports are challenging for existent HME models and athletes are measured. Their performance is currently not good enough to use them in performance assessments of athletes, which we try to change. 
Our contributions can be summarized as follows:
\begin{itemize}[leftmargin=0.5cm,itemsep=0pt, topsep=-0.5\parskip]
\item We reveal inconsistencies in the GT data of ASPset \citep{aspset} and fit3D \citep{fit3d}. The body shape of a single person varies mistakenly in the GT.

\item We create and evaluate different models to convert between anthropometric measurements and SMPL-X body shape parameters for all genders. 
We call them \emph{A2B}. 

\item We analyze and compare the performance of existing HME models on ASPset and fit3D. Replacing the estimated body shape parameters (and keeping the pose) of each HME model with A2B body shape parameters increases the performance of all models.

\item With fine-tuned SOTA 2D and 3D HPE models \citep{vitpose, uu}, IK, anthropometric measurements, and our A2B model, we estimate accurate human meshes with a consistent body shape. We show that this approach achieves superior results to SOTA HME models, although still evaluated on the inconsistent GT. 


\item Our models and code for our approach are publicly available: \url{https://github.com/kaulquappe23/a2b_human_mesh}
\end{itemize}

%% file: sec/2_related_work.tex
\section{Related Work}

Human Mesh Estimation (HME) is an active area of research. Body models like SMPL \citep{smpl} and its successor SMPL-X \citep{smplx} are broadly used. Their advantage is that they decouple human pose and shape. The pose parameters $\theta$ give the rotations of the joints relative to the parent joint. The shape parameters $\beta$ model the basic body shape. At first, a mesh is created with a linear mapping from $\beta$ parameters to a T-shaped pose. Next, some pose-specific shape deformations are applied, and then the mesh is rotated at the joints according to the $\theta$ parameters.

The first HME model that estimates SMPL-X meshes from images, SMPLify-X, was introduced along SMPL-X \citep{smplx}. It detects 2D image features and then fits an SMPL-X model to these. To achieve that, they incorporate a pose prior trained on a large motion capture dataset and an interpenetration test. A more recent model for HME is Multi-HMR \citep{multihmr}. It predicts 2D heatmaps for person centers and based on that the human mesh with a human prediction head. OSX \citep{osx} is a HME model using a component aware Transformer that is composed of a global body encoder and local decoders for face and hands. SMPLer-X \citep{smplerx} is introduced as a generalist foundation model for HME trained on a large amount of datasets mainly using vision transformers. There are many other HME models, some focussing more on whole-body HME \citep{expose, pixie, hand4whole}, others on multi-person HME - either with a two stage approach using a person detector and a single person human mesh estimator \citep{3DCrowdNet, hmr2, GRM3D}, or a single stage approach estimating the meshes of all persons at once \citep{psvt, romp, pymaf}.

\citet{shapy} observed that existing HME models focus more on the body pose than the shape, although the shape is equally important for many applications. They propose SHAPY, a model that uses anthropometric and linguistic attributes to create accurate body shapes. Moreover, \citet{soy} introduce SoY, which contains specific loss functions to enhance the body shape accuracy. 
AnthroNet \citep{anthronet} propose a new body model that is learned with an end-to-end trainable pipeline. It takes anthropometric measurements as an input to learn a mesh model that accurately captures shapes of humans, but this model is different from the commonly used SMPL-X model. We use the common SMPL-X body model and decouple the estimation of the shape from the estimation of the pose. Sengupta et al. \cite{sengupta} estimate anthropometric measurements from images and use a linear layer to convert them to body shape parameters. However, their model operates on single images and their measurement to shape conversion is different from ours and only available for a single gender. We further ensure the consistency of the body shape over time.

In the last years, 3D HME approaches leveraged Inverse kinematics (IK) to enhance their results. HybrIK \citep{hybrik} transforms 3D joint coordinates to relative body-part rotations for 3D HME by using a twist-and-swing decomposition. HybrIK-X \citep{hybrikx} further enhances HybrIK with expressive face and hands. 
Cha et al. \citep{3dhmeik} leverage IK to tackle the challenge of person-to-person occlusions in images with interacting persons. PLIKS \citep{pliks} (Pseudo-Linear Inverse Kinematic Solver) approaches HME by analytically reconstructing the human model via 2D pixel-aligned vertices in an IK-like manner. 

Although HME is an active area of research, it is yet not common in computer vision for sports. Due to high velocities and a great variation of poses, sports is a challenging scenario for all kinds of human pose and shape estimation. The fit3D dataset \citep{fit3d} is a dataset which consists of videos from gym sports exercises with repetitions and is annotated with human meshes. AIFit \citep{fit3d} is a tool trained on fit3D which can reconstruct 3D human poses, reliably segment exercise repetitions, and identify the deviations between standards learned from trainers, and the execution of a trainee. Other sports datasets only consist of 3D joint annotations, like ASPset \citep{aspset} or SportsPose \citep{sportspose}. SportsCap \citep{sportscap} is an approach for simultaneously capturing 3D human motions and understanding fine-grained actions from monocular challenging sports videos. 

%% file: sec/3_3d_gt_errors.tex
\section{Errors in 3D Human Shape Ground Truth}\label{sec:gt_flaws}

Each person has a specific basic body shape that does not change over a short time period. Therefore, the SMPL-X body model decouples the human pose encoded by $\theta$ parameters from the basic body shape encoded by $\beta$ parameters. Deformations to the basic body shape that are caused by the current pose are modeled separately. Therefore, it makes sense to assign a single set of shape parameters $\beta$ to a person for a given short time period such as a recorded action to describe his/her shape. Further, there are lengths that can be calculated from 3D joints that should never change, since individual bones of humans are rigid and should not be deformed by different poses. Our approach enforces a single set of shape parameters per person and immutable bone lengths.

As a first step, we analyze if the GT data of our used datasets fulfills these properties. 
\begin{table}[tb]
\centering
\resizebox{\linewidth}{!}{ 
\begin{tabular}{c|ccc||c|ccc}
\toprule
\multicolumn{4}{c||}{ASPset} & \multicolumn{4}{c}{fit3D}\\
Measure & $\sigma$ & r. $\sigma$& r. range & Measure & $\sigma$ & r. $\sigma$ & r. range\\
\midrule
head & 0.91 & 5.98\% & 57.91\%              	&head & 0.73 & 2.73\% & 17.52\%\\
hip width & 1.71 & 9.48\% & 85.46\% 			&hip circ. & 0.87 & 0.84\% & 8.17 \%\\
forearm & 1.99 & 8.37\% & 92.04\% 			&forearm & 0.34 & 1.40\% & 9.24\%\\
upper arm & 1.72 & 6.29\% & 66.35\% 		&arm & 0.76 & 1.51\% & 9.42\%\\
lower leg & 1.44 & 3.60\% & 41.36\%			&lower leg & 0.52 & 1.31\% & 13.80\%\\
thigh & 1.65 & 4.23 \% & 35.46\%				&thigh & 0.43 & 1.17\% & 11.74\%\\
 & & &     													&height & 1.60 & 0.94\% & 8.69\% \\ 
  & & &     													&$\beta$ param. & 0.64 \\
  \bottomrule
\end{tabular}
}
\caption{GT data analysis for ASPset (left) and fit3D (right):  Standard deviation $\sigma$, relative standard deviation $\frac{\sigma}{\textit{avg}}$ and relative range $\frac{\max - \min}{\textit{avg}}$ of anthropometric measurements. Standard deviations are given in cm, but not for the $\beta$ parameters. The values are averaged between left and right body parts and between all persons used for evaluations in Section \ref{sec:5_a2b_hme}. The $\beta$ parameter standard deviation is averaged over all $\beta$ parameters.} \label{tab:gt_analysis}
\vspace{-0.4cm}
\end{table}
In this paper, we use ASPset \citep{aspset} and fit3D \citep{fit3d}, since both datasets consist of videos with fast changing poses and 3D GT. Results for the Human3.6M \citep{h36m} and MPI-INF-3DHP \citep{mpi_inf_3d} datasets are presented in the supplementary. For ASPset, we analyze bone lengths, since it has only GT annotations for 3D joints. For fit3D, GT SMPL-X $\beta$ parameters are available, hence we can analyze the $\beta$ parameters directly and further the derived anthropometric measurements. These values are the output of our deterministic B2A function: It generates a standard T-pose with the given $\beta$ parameters and computes 36 anthropometric measurements from the resulting mesh. Results of our GT analysis for a subset of the anthropometric values are shown in Table \ref{tab:gt_analysis}. We can see that the GT itself is not consistent. The deviations are larger for ASPset. Although we have GT SMPL-X meshes for fit3D, every $\beta$ parameter of a single person has a standard deviation of 0.64 on average.\footnote{Averaged standard deviation means (in the whole paper) that the standard deviation is calculated per person, and the mean of the resulting standard deviations is calculated afterwards.} This is a relevant flaw in the GT shape annotation, since based on the model, the GT shape should be consistent for each human. Nevertheless, we use the given inconsistent GT for our evaluations for comparability with related work and as we have no good means to correct them. The reader should keep this in mind. Nevertheless, we want to encourage future research in the field of 3D human pose and mesh data collection to try to eliminate these flaws in the provided GT.

%% file: sec/4_a2b.tex
\section{From Measurements to Body Shape}\label{sec:A2B}

Humans have been measured for centuries \citep{tailor}. Tailors know exactly which measurements to take for perfectly fitting a suit or dress to the body shape of a customer. In sports, it is already common practice that professional athletes are measured for precise performance assessments. Measuring a human is easy and well understood. In contrast, the parameters of the body shape for human mesh models like SMPL-X \citep{smplx} are not humanly interpretable. The $\beta$ parameters describe the principal components of the human body shape with typically around 10 to 16 values and are the result of a PCA executed on the human meshes of a training dataset while learning the SMPL-X model. Fixing all $\beta$ parameters despite one and looking at the results lets human observers get a notion of what this parameter might mean, but in total, the $\beta$ parameters and their interactions are not well interpretable. Therefore, we want to leverage the well established technique of measuring humans to create precise body shape parameters for the commonly used SMPL-X human mesh model. We call our approach to convert from 36 \underline{A}nthropometric measurements \underline{to} \underline{B}ody shape parameters \emph{A2B}. Since there is no known relation between anthropometric measurements (AMs) and $\beta$ parameters, our aim is to learn this mapping. The reverse direction, \emph{B2A}, is a deterministic function of the human mesh, as the AMs can be measured from the mesh. 

\subsection{Data Generation}\label{sec:datagen}

We select 36 anthropometric measurements for our models based on the selections of AnthroNet \citep{anthronet} and an anthropometry study of the U.S. army \citep{usarmy}. They can be categorized into 23 lengths and 13 circumferences. Apart from the bone lenghts like arm length, thigh length, etc., this includes also detailed measurements like shoulder width, front torso height, lateral neck length, waist circumference, calf circumference, etc. A visualization and precise description of all AMs can be found in the supplementary.

Many existing datasets provide a wide range of different poses, but most incorporate the same humans. For learning a conversion model from AMs to $\beta$ parameters, we need a lot of samples for different humans, no matter the pose. With given shape parameters, we can use the B2A function to compute the AMs. Recall, B2A is a deterministic function measuring the AMs from meshes in T-pose.

\begin{figure}[tb]
  \centering
    \includegraphics[width=\linewidth]{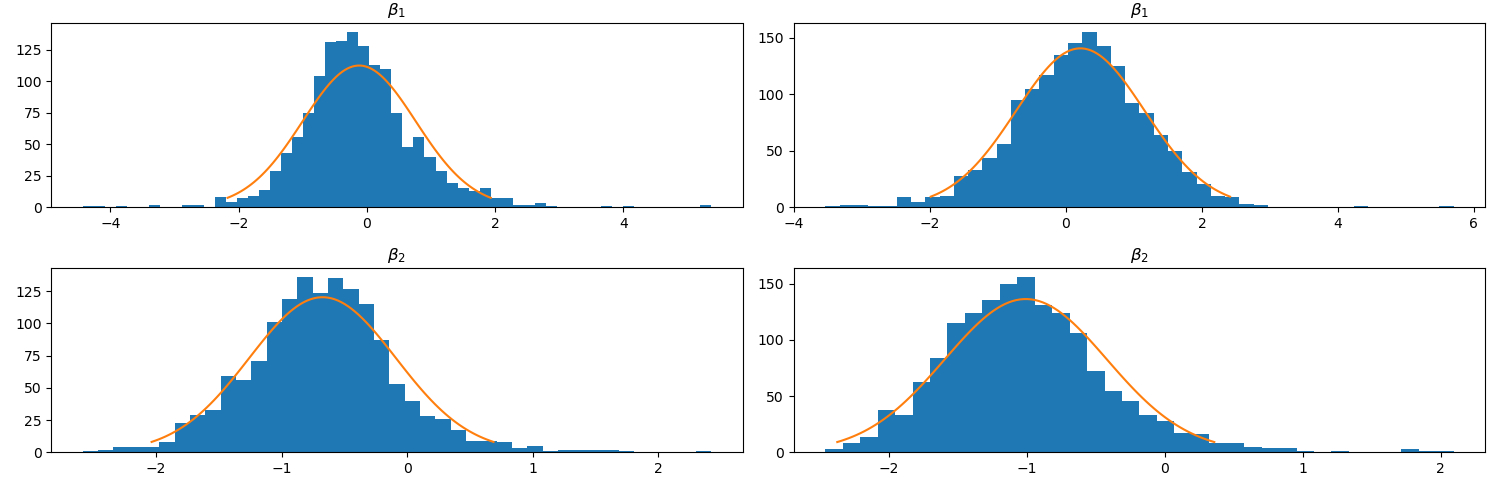}
   \vspace{-0.5cm}
   \caption{Histograms and fitted normal distribution (orange) for the first two $\beta$ parameters for all male (left) and female (right) subjects of the AGORA \citep{agora} dataset.}
   \label{fig:histograms}
   \vspace{-0.4cm}
\end{figure}

Because many different body shapes are required for the learning process, we use the AGORA \citep{agora} dataset. It consists of 1447 male and 1588 female subjects. We are not able to use the larger dataset from AnthroNet \citep{anthronet}, since it uses its own mesh model and the authors did not publish their conversion to the SMPL-X model, which we want to use as it is most commonly used in research. Although comparably large, 1447/1588 subjects is still a little amount of data to learn a model. Hence, we analyze the $\beta$ parameters in the AGORA dataset with the aim to randomly sample more data with realistic body shapes. Histograms (see Figure \ref{fig:histograms}) of the occurring $\beta$ parameters show that their distribution roughly follows a normal distribution. Therefore, we train our models with randomly sampled data according to these distributions, either assuming a normal distribution fitted to the histograms or a uniform distribution with the same minimum and maximum values as in the data analysis. This means that we sample each $\beta$ parameter according to the selected distribution, create the mesh according to the sampled values and derive the AMs with B2A. With this strategy, we can create a dataset with as many subjects as we need.
As we do not expect the analyzed AGORA data to cover the full range of human body shapes, we also train with extended distributions, meaning that we increase the standard deviation $\sigma$ to $\alpha_n\sigma$ in the case of a normal distribution or stretch the interval by a factor $\alpha_u$ in case of a uniform distribution. 

\subsection{Models}

We use the same number of $\beta$ parameters for each gender as used in the AGORA dataset, meaning $11$ for male, $10$ for female, and $16$ for neutral subjects. With $36$ AMs as input values and $10-16$ output values for our A2B models, the dimensionality of the data is low. Therefore, we experiment with Support Vector Regression (SVR) and with small neural networks (NN). We split the AGORA dataset in an 80\% train, 15\% test and 5\% validation subset. For SVR, we additionally randomly sample $10,000$ subjects for training. We use a hyperparameter search based on the validation split to determine the optimal settings, which leads us to a radial basis function kernel, an error margin of $\epsilon = 0.012$ and a regularization constant of $C=3791$. For the NNs, we randomly sample new data in each iteration. The hyperparameter search for the NNs results in a model with $4$ layers, $330$ neurons per layer, tanh as activation function, and Xavier Glorot as initialization. We use mean squared error on the model output (the $\beta$ parameters) as training loss.

\subsection{Results}

\begin{table}[tb]
\centering
\resizebox{\linewidth}{!}{ 
\begin{tabular}{cc|ccc|ccc}
\toprule
& & \multicolumn{3}{c|}{$\diameter$ error of $\beta$} & \multicolumn{3}{c}{$\diameter$ error of $A$ [in mm]}\\
 & train data & m & f & n & m & f & n\\
\midrule
NN & AGORA & 			9.11 & 13.9 &  24.0 & 0.814 & 0.934 & 1.459\\
NN & norm. & 			2.62 & 4.34  & 18.0 & 0.356 & 0.392 & 1.711\\
NN & norm. ext. & 	1.87 & 3.69& {14.8} & 0.248 & 0.285 & {1.384}\\
NN & unif. & 			5.08 & \textbf{1.25} & \textbf{2.81} & \textbf{0.243} & \textbf{0.268} & 0.623\\
NN & unif. ext. &  \textbf{1.61} & 3.20 & {8.37} & 0.274 & 0.419 & \textbf{0.381}\\
\midrule
SVR & AGORA & 		2.56 &  16.1 & 3.82 &  1.659 & 5.195 &  2.557 \\
SVR & norm. & 			4.08 &  17.8 &  59.0 & 2.975 & 4.303 & 14.63\\
SVR & norm. ext. & 	0.210 & 4.60 & 6.27 & 0.280 & 1.090 & 2.211\\
SVR & unif. & 		0.0396 & 0.0350 & \textbf{0.162} & 0.124 & 0.284 &  0.214\\
SVR & unif. ext. & \textbf{0.0252} & \textbf{0.0193} & 0.306 & \textbf{0.082} & \textbf{0.136} & \textbf{0.164}\\
 \bottomrule
\end{tabular}
}
\caption{Results of our A2B models on the test split of the AGORA dataset. The first block ($\beta$) shows the error if we take the GT $\beta$ parameters, derive 36 anthropometric measurements (B2A), input them into the A2B models and evaluate the MSE of the predicted $\beta$ parameters in the scale $10^{-3}$. The second block ($A$) calculates B2A from the predicted $\beta$ parameters and evaluates the mean difference between the GT and predicted AMs (all 36) in mm. Results are given for m(ale), f(emale), and n(eutral) models. Further visualizations are in the supplementary.} \label{tab:atb_res}
\vspace{-0.3cm}
\end{table}
We train each model (NN and SVR) for each gender and with different dataset variants:  We train solely on the AGORA train split, as well as on uniformly and normally distributed randomly sampled data according to the data analysis, and we further extend the range of the data as described in Section \ref{sec:datagen} with $\alpha_n = \alpha_u = 1.5$. 
The results are displayed in Table \ref{tab:atb_res}. We evaluate the performance of our models in two ways. At first, we calculate the error of the predicted and GT $\beta$ parameters. Second, we calculate the mean deviation of the AMs of the meshes from the predicted and GT $\beta$ parameters ($A$). Therefore, this evaluation can further be seen as a kind of cycle consistency evaluation of A2B (our learned model) and B2A (the deterministic measuring function). We provide a visualization of the evaluation process and evaluation results of the real-world SSP-3D dataset \cite{ssp3d} in the supplementary.
The anthropometric error is our main metric as 
these values reflect the desired body shape given as an input by the user and are further interpretable. The $\beta$ parameters are somehow arbitrary in their scale. For all genders and SVR, using an extended uniformly sampled dataset works best. For the NNs, a uniformly sampled dataset works best for male (m) and female (f) genders and an extended normally sampled dataset for the neutral (n) meshes. The results for the neutral model are worse in general, especially in the case of the NNs, which might be due to the fact that the neutral model needs to express a more diverse range of body shapes. Furthermore, the SVR achieves better results for all genders.
Thus, we use these models for all datasets, without any fine-tuning or adaptation to specific datasets.

%% file: sec/5_a2b_hme.tex
\section{Leveraging A2B Model Results for HME}\label{sec:5_a2b_hme}

Now that we have trained the A2B models, we can use them to generate precise body shape parameters upfront and reuse them for every evaluation of a specific person. In the next section, we describe how the A2B results can be used to improve existing HME models (see Section \ref{sec:hme_res}). Further, \textbf{we introduce a new approach to HME} (see Section \ref{sec:a2b_hme}). We leverage the good performance of a sequence-based 2D-to-3D uplifting HPE model and convert the 3D stick-figure poses to human meshes with the help of our A2B models. With this approach, we achieve superior results compared to existing HME models. However, we want to emphasize that our approach is not exactly comparable to existing ones since it uses the additional information of anthropometric measurements. Since the performance of existing HME approaches is not good enough to be used for sports analyses, our \textbf{main goal is to achieve the best possible performance with marginal additional information}. As professional athletes are measured anyway, this results in actual no overhead in these use cases.

We evaluate all models on the ASPset \citep{aspset} 3D human pose dataset. It consists of various different sports motion clips performed by different subjects, recorded from three camera perspectives. We evaluate on the test set, which contains two subjects and 30 videos for each subject. In the test set, only one camera perspective is public, so we evaluate on this perspective. Evaluating SMPL-X meshes for ASPset is non-trivial. Regressing standard SMPL-X joints from SMPL-X meshes is built-in, but for all other keypoint definitions it is necessary to define a custom regressor. Since there is no regressor available for ASPset, we create a custom SMPL-X regressor \citep{regressor}. 
 
We further evaluate on fit3D \citep{fit3d}, since this is the only sports dataset with public SMPL-X annotations. We evaluate the meshes and the SMPL-X joints since they are available. We select a subset of 37 SMPL-X joints. Since our focus is mainly on the body and not on the hands and face, we remove a lot of these joints and consider only the main body pose for MVE calculation. A list of the selected joints can be found in the supplementary.  Hence, we achieve a fair comparison with this evaluation scheme. For both datasets, we do not have access to the athletes to measure them. 
Therefore, we simulate athlete measuring by measuring the GT meshes.
Details can be found in the supplementary. Since there is no GT available for the official test set evaluation on the evaluation server of fit3D, we split the official training dataset into a training, validation, and test set for our evaluations. We perform a leave-one-out cross validation and average the results.

Sports datasets differ from most commonly used everyday activity datasets in the aspect that the poses are more diverse and the motions are faster, which makes sports datasets more difficult. In some cases, the poses are so difficult that some models do not detect a human at all. This makes a fair evaluation hard, since the standard MPJPE metric takes the mean of the joint position errors. Assuming a default pose for all frames where no person is detected would result in a very high error that shifts the mean enormously. Hence, we report the MPJPE only on the frames where persons are detected. Since mostly difficult frames are omitted, this will result in a slightly easier setting for methods that find fewer persons, but we include the number of missing frames in our results for comparison. 

\subsection{Improving HME Model Results}\label{sec:hme_res}

A major problem for HME based analyses is a varying basic body shape within a single video. Existing HME models output different $\beta$ parameters for each frame. Exemplarily, we show the standard deviation of the body height of one subject in Table \ref{tab:other_models_res}. Recall that these measurements and $\beta$ parameters are based on a T-pose mesh, hence varying poses have no influence on measuring and $\beta$ parameters. Using $\beta$ parameters generated with A2B models solves this problem. 
The necessary 36 measurements are either measured from the human directly, or averaged from the provided GT mesh (fit3D) or IK applied to the GT poses (ASPset). We call these measurements pseudo GT and include more details in the supplementary. We choose this process to simulate real measurements which exist for most professional athletes. 
We combine existing HME models with the body shape estimated by our A2B models by replacing the estimated $\beta$ parameters with the ones predicted by the A2B models. 
We select three recent well performing models on the AGORA dataset (SMPLer-X \citep{smplerx}, OSX \citep{osx}, Multi-HMR \citep{multihmr}), and the first HME method developed by the SMPL-X authors, SMPLify \citep{smplx}. Since SMPLer-X is trained on the official training data of fit3D, an evaluation with this model is not meaningful, and we omit it here. Moreover, SMPLify-X is not SOTA anymore and achieved the worst results for ASPset. Therefore we omit it, too. 

The first evaluation contains the original result from the respective model, and evaluations where the pose from the model is kept, but the $\beta$ parameters are replaced with the A2B body shape parameters with pseudo GT input. 
Results are displayed in Table \ref{tab:other_models_res}. The results for the MVE of the meshes for fit3D are included in Table \ref{tab:ik}. We can see that for all models, replacing the estimated $\beta$ parameters by $\beta$ parameters from our A2B models with pseudo GT input leads to an improvement. For one model, the gendered meshes outperform the neutral ones and for all other models, the neutral meshes perform best. We use the correct gender (male or female) of the subject in the gendered results. Interestingly, the NN outperforms the SVR for all neutral experiments, although the SVRs achieved better results on the AGORA dataset evaluation. The reason could be that AGORA is a synthetic dataset and does not reflect reality. SMPLer-X achieves the best results for ASPset and Multi-HMR for fit3D, both with a significant margin. OSX performs worse on fit3D than on ASPset, but Multi-HMR performs better by a large margin and surpasses OSX. All methods benefit from our A2B $\beta$ parameters based on pseudo GT with MPJPE improvements from 11\,mm to 3\,mm regarding both datasets and MVE improvements of approx. 8\,mm regarding fit3D. 
\begin{table}[tb]
  \centering
\resizebox{\linewidth}{!}{ 
\begin{tabular}{cc|cc|cccc|c}
\toprule
& Model & orig. & $\sigma$ & NN g & SVR g & NN n & SVR n & no r. $\downarrow$\\
\midrule
\parbox[t]{2.3mm}{\multirow{4}{*}{\rotatebox[origin=c]{90}{ASPset}}}
&SMPLer-X & 86.0 & 2.9 & 78.9 & 78.5 & \textbf{78.3} & 78.5 & 0.11\% \\
&OSX & 92.3 & 0.2 & 89.6 & \textbf{89.3} & 89.4 & 89.6 & 0.10\% \\
&Multi-HMR & 102.5 & 3.6 & 100.0 & 100.3 & \textbf{99.3} & 99.5 & 0.44\% \\
&SMPLify-X & 138.2 & 13.0 & 127.7 & 127.4 & \textbf{126.8} & 126.9 & 0.02\% \\
\midrule
\parbox[t]{2.3mm}{\multirow{2}{*}{\rotatebox[origin=c]{90}{fit3D}}}
&OSX & 94.2 & 3.9 & 88.9 & 88.6 & \textbf{87.1} & 87.2 & 3.45\% \\
&Multi-HMR & 74.6 & 3.3 & 69.6 & 69.6 & \textbf{68.0} & 68.4 & 1.54\% \\
 \bottomrule
\end{tabular}
}
\caption{MPJPE results in mm for existing models on the test splits of ASPset (top) and fit3D (bottom). The second column (\emph{orig.}) contains the original results, the other columns results with replaced $\beta$ parameters from our \textbf{A2B models with pseudo GT anthropometric measurements} as input and either gendered (g) or neutral (n) meshes, and the percentage of frames with no result (no r.). The $\sigma$ column displays the mean standard deviation of the body height per subject in cm for the original results, while all A2B body shapes have $\sigma = 0$.} \label{tab:other_models_res}
\vspace{-0.2cm}
\end{table}
\begin{table}[tb]
\centering
\resizebox{\linewidth}{!}{ 
\begin{tabular}{cc|c|c|cccc}
\toprule
&Model & orig. & median & NN g & SVR g & NN n & SVR n \\
\midrule
\parbox[t]{2.3mm}{\multirow{4}{*}{\rotatebox[origin=c]{90}{ASPset}}}
&SMPLer-X & 86.0 & 86.0 & 85.9 & \textbf{85.7} & 86.0 & 86.0\\
&OSX & 92.3 & 92.4 & 92.4 & \textbf{92.2} & 92.3 & 92.4\\
&Multi-HMR & 102.5 & \textbf{102.0} & 102.6 & 103.0 & \textbf{102.1} & 102.2\\
&SMPLify-X & 138.2 & 133.6 & 133.8 & \textbf{133.5} & 133.6 & 133.5\\
\midrule
\parbox[t]{2.3mm}{\multirow{2}{*}{\rotatebox[origin=c]{90}{fit3D}}}
&OSX & 94.2 & \textbf{93.0} & 95.0 & 94.8 & \textbf{93.0}  & \textbf{93.0} \\
&Multi-HMR & 74.6 & \textbf{73.9} & 75.8 & 76.1 & \textbf{73.9} & 74.1\\
 \bottomrule
\end{tabular}
}
\caption{MPJPE results in mm for existing models on the test split of the ASPset (top) and fit3D (bottom) datasets. The second column contains the original results, the other columns results with replaced $\beta$ parameters. Either the median $\beta$ parameters are used or the results from our \textbf{A2B models with median anthropometric measurements from the respective model} as input. 
} \label{tab:other_models_betas}
\vspace{-0.6cm}
\end{table}

Although it is not our main goal, we further evaluate the capabilities of a fixed body shape without available GT measurements to ensure consistent body shapes in the case that no measurements are available. The simplest approach is to use the median of the $\beta$ parameters across all frames of the respective model. However, the $\beta$ parameters have no real meaning. Therefore, we compare this approach to taking the median of the anthropometric measurements of the generated meshes and then converting them to $\beta$ parameters via the A2B models.
Results are displayed in Table \ref{tab:other_models_betas}. 
For SMPLer-X and OSX, using the median $\beta$ parameters lead to equal or even worse results on ASPset. Regarding ASPset, using our A2B models increases the performance of all models slightly. Switching from the neutral output that these models all have to a gendered model works best in most of these cases, but the neutral A2B models also lead to a marginal improvement. Regarding fit3D, using the median $\beta$ parameters already enhances the MPJPE and MVE results. Using $\beta$ parameters from an A2B model leads to the same improvement for both metrics, OSX achieves the best results with SVR and the neutral model, Multi-HMR with NN and the neutral model.

\begin{figure*}[tb]
  \centering
    \includegraphics[width=0.95\linewidth]{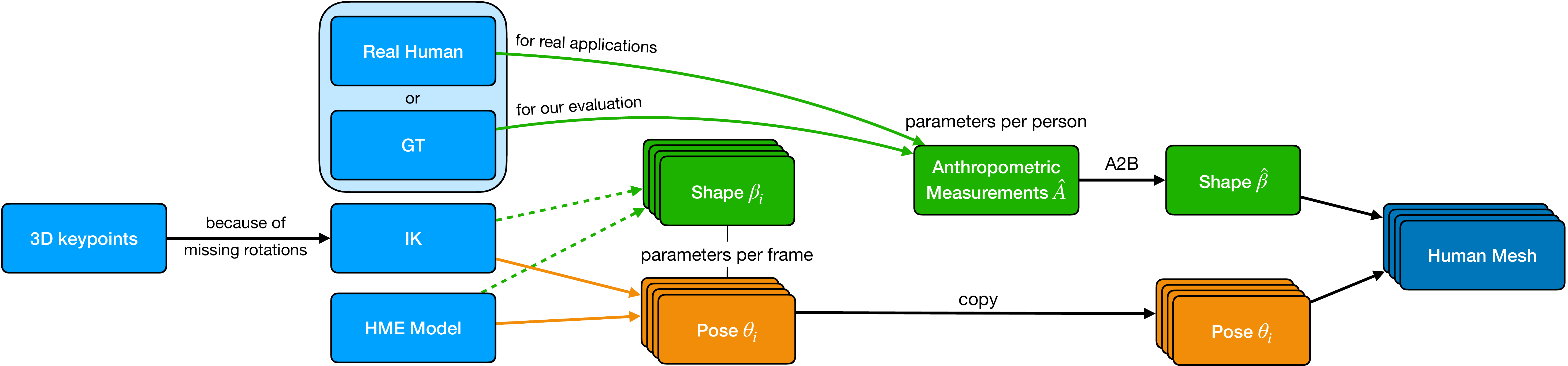}
    \vspace{-0.1cm}
   \caption{Overview of our inference pipeline. The pose and shape parameters are obtained either from IK applied to UU results (Sec. \ref{sec:ik_res}) or from an HME model (Sec. \ref{sec:hme_res}). In real applications, the anthropometric measurements will be taken directly from the humans. For our evaluations, we use the GT shape parameters and further experiment with the shape parameters of the respective model (IK or HME).}
   \label{fig:pipeline}
   \vspace{-0.3cm}
\end{figure*}
Anthropometric measurements can further be used to easily convert between neutral and gendered (male or female) models. In contrast, $\beta$ parameters are not transferable between models of different genders. Therefore, until now, the conversion could only be achieved by minimizing the MVE between meshes of different genders in an iterative process. We can now use the B2A function to obtain measurements for a mesh of one gender and apply the A2B model of the other gender to these anthropometric measurements in order to get the corresponding $\beta$ parameters for this gender.

\subsection{HME with Sequence Based 3D HPE and A2B}\label{sec:a2b_hme}

All evaluated HME models are working image-wise. In contrast, SOTA 3D HPE models take a long sequence of 2D poses as an input, which helps to capture movements precisely. The models are called uplifting models, since they lift 2D pose sequences to 3D pose sequences. We use the efficient SOTA 3D HPE model \emph{uplift and upsample} (UU) \citep{uu} to estimate the 3D poses on videos. To estimate the required 2D poses from the video frames, we use ViTPose \citep{vitpose}, a SOTA 2D pose estimation model. It is important to note that UU operates on pose sequences instead of single frames like the HME models in Section \ref{sec:hme_res} and can leverage the information of neighboring frames to estimate a more sophisticated pose. Since we have GT 3D joints available, we can fine-tune the models (ViTpose for 2D HPE and UU for 3D HPE) on our data. This is also necessary to adapt the model to the dataset specific joint definitions since many 3D HPE models like UU are pretrained on datasets like Human3.6M \citep{h36m}, but those joint definitions do not match ASPset nor fit3D. We fine-tune both 2D and 3D HPE models on the training subsets. On the test subsets, UU achieves an MPJPE of 63.85\,mm on ASPset and an MPJPE of 29.60\,mm on fit3D, which is better than the best existing HME model for both datasets (see Section \ref{sec:hme_res}). However, UU only outputs 3D joints, no meshes. Moreover, a stick-figure 3D pose is not sufficient to model the pose parameters $\theta$ of the SMPL-X mesh, since some rotations are missing. Hence, it is impossible to calculate the necessary rotation parameters directly from the UU result.

\subsubsection{Inverse Kinematics for Full Pose Estimation}\label{sec:ik}

Because we need the rotation parameters, we use the well established approach of inverse kinematics (IK) with a pose prior to obtain the missing rotations by fitting an SMPL-X mesh to the 3D joint locations estimated by UU. IK outputs the best SMPL-X parameters ($\beta$ and $\theta$) that fit the mesh to the given 3D joint locations. Details can be found in the supplementary.

\subsubsection{Experiments and Results}\label{sec:ik_res}

We evaluate different experiments in Table \ref{tab:ik}. For comparison, we mention the UU 3D HPE performance (first rows for each dataset in Tab. \ref{tab:ik}). These results correspond to stick-figure poses and not the required meshes. Therefore, they are not directly comparable to the other results. 

Our main approach is shown in the second rows in Table \ref{tab:ik}. We evaluate the results of IK applied to the UU joint locations with original, median, and pseudo GT based A2B $\beta$ parameters. The real-world scenario corresponds to the following. GT measurements can be measured from the athlete directly and the 3D pose can be estimated with UU and IK. The $\beta$ parameters are estimated with the A2B models. A visualization of this pipeline can be found in Figure \ref{fig:pipeline}.
We include the best result(s) from existing HME models in the respective last rows for comparison and provide qualitative results in Figure \ref{fig:vis}. \textbf{Our approach outperforms the best existing HME model for both datasets by a large margin.} But we need to mention that our approach is not directly comparable to the original results of the HME models, since our approach needs the additional information of measurements, but they already exist in our scenario. However, our approach still outperforms the existing HME models even when they use the same measurements and A2B results as our model does. Further, our model provides results for all frames, which is not the case for the other HME models.

  \begin{table*}[tb]
    \centering
    \resizebox{0.85\linewidth}{!}{ 
    \begin{tabular}{cccc|cccccccc}
    \toprule
    & \multicolumn{3}{c|}{\emph{inconsistent shape}} &  \multicolumn{8}{c}{\emph{consistent shape (ours)}}\\
    DS & pose & orig. & $\sigma$ & measure & NN g & SVR g & NN n & SVR n & median & $\sigma$ &  no r. $\downarrow$ \\
    \midrule
    \parbox[t]{2.3mm}{\multirow{4}{*}{\rotatebox[origin=c]{90}{ASPset}}} 
    & UU & 63.9 & & no mesh\\
    & IK-UU  & {67.5} & 3.0&  GT & {56.4} & {56.6} & {\textbf{55.2}} & \textbf{55.2} & {-} & 0.0 & 0.0\%\\
    & IK-UU & {67.5} & 3.0&  IK-UU & {66.9} & {\underline{66.6}} & {67.3} & {67.1} & {67.2} & 0.0 & 0.0\% \\
    &SMPLer-X & {86.0} & 2.9 & GT & 78.9 & 78.5 & \underline{78.3} & 78.5& - & 0.0 & 0.11\%\\
    \midrule
    \parbox[t]{2.3mm}{\multirow{7}{*}{\rotatebox[origin=c]{90}{fit3D}}} 
    & UU & 34.3 & - & no mesh\\
    & IK-UU    & 38.5 / 46.3 & 8.7 & GT        & 41.2 / 47.5 & 41.3 / 46.9             & 38.8 / \textbf{45.3}    & \textbf{38.7} / \textbf{45.3}       & - & 0.0 & 0.0\%\\
    & IK-UU    & 38.5 / 46.3 & 8.7 & IK-UU     & 42.6 / 51.2 & 41.6 / 48.6             & \underline{39.8} / \underline{47.8}             & \underline{39.8} / \underline{47.8} & 39.9 / \underline{47.8} & 0.0 & 0.0\%\\
    &Multi-HMR & 74.6 / 76.1 & 3.3 & GT        & 69.6 / 67.8 & 69.6 / \underline{67.6} & \underline{68.0} / 68.0 & 68.4 / 68.8                         & - & 0.0 & 1.54\%\\
    &Multi-HMR & 74.6 / 76.1 & 3.3 & Multi-HMR & 75.8 / 77.3 & 76.1 / 76.7             & \underline{73.9} / 75.6 & 74.1 / 75.8                         & \underline{73.9} / \underline{75.5} & 0.0 & 1.54\%\\
    &      OSX & 94.2 / 89.0 & 3.9 & GT        & 88.9 / 83.4 & 88.6 / 82.3             & \underline{87.1} / \underline{81.1} & 87.2 / \underline{81.1}             & - & 0.0 & 3.45\%\\
    &      OSX & 94.2 / 89.0 & 3.9 & OSX       & 95.0 / 91.7 & 94.8 / 90.2             & \underline{93.0} / 87.7             & \underline{93.0} / \underline{87.6} & \underline{93.0} / \underline{87.6} & 0.0 & 3.45\%\\
     \bottomrule
    \end{tabular}
    }
    \vspace{-0.1cm}
    \caption{MPJPE and MVE results in mm on the test splits of ASPset (top) and fit3D (bottom) of our approach compared to the respective best HME model(s). For fit3D, we calculate the MVE, since we have GT meshes available. We display it as the second value in every column. The \emph{pose} column indicates the origin of the pose. The \emph{orig} column contains the result as it is estimated from the method indicated in the \emph{pose} column (with inconsistent body shapes). The right block contains the results with the originally estimated $\beta$ parameters replaced by consistent ones. The \emph{measurements} column indicates which anthropometric measurements are used for the A2B computation (which $\beta$ parameters are used for the median computation) whose results are the replacement $\beta$ parameters in the last five columns. We highlight the overall best results for \emph{estimated} meshes with \emph{consistent shapes} in bold and underline the best (MPJPE and MVE) results in each line. We further add the mean standard deviation of the body height and the percentage of frames with no result as in Table \ref{tab:other_models_res}.}
    \label{tab:ik}
    \vspace{-0.4cm}
    \end{table*}

We analyze the results of the building blocks of our model in detail. Applying IK to the UU results deteriorates the UU results by nearly 4\,mm for ASPset and 5\,mm for fit3D (see Tab. \ref{tab:ik}, first and second rows, column \emph{orig.}), but this step is necessary since the UU result is only a stick-figure pose and not sufficient for our purpose. Moreover, these results are still better than the best existing HME model (see last rows in Tab. \ref{tab:ik}). 

Next, we replace the inconsistent $\beta$ parameters with the results from our A2B models. This is especially helpful for our approach since IK produces body shapes with high inconsistencies, as shown by the larger standard deviation of the body height compared to other HME models.
For ASPset, using pseudo GT AMs results in a large improvement of over 12\,mm. Remarkably, this result surpasses even the original UU result by 8\,mm. It seems that incorporating a clearly defined mesh helps to fix some typical errors of UU and enhance its result in case of ASPset. 
In general, the error on fit3D is much lower for UU based approaches. The reason might be that it consists of much more data, such that we can fine-tune UU for a longer time. Further, the videos are recorded in a lab in comparison to the in-the-wild videos of ASPset. The lab environment is very similar to the Human3.6M dataset \cite{h36m}, which serves as a training dataset for most recent HME models. Therefore, the results of ASPset are more relevant for future applications of our approach, where we assume only a few available 3D annotations and in-the-wild recordings. For fit3D, applying the A2B body shapes from pseudo GT AMs leads to a slight decrease in performance of 0.2\,mm. Inconsistent shapes in the GT (see Section \ref{sec:gt_flaws}) are likely to cause this behavior. Still, our approach using a 3D HPE model and IK outperforms all existing HME models, no matter if the original inconsistent or the consistent body shapes from A2B are used. 

Regarding the gendered meshes, we observe that the performance is slightly better for male than for female subjects. fit3D consists of two female and six male subjects. The best score of 40.2\,mm for the male subjects is achieved with the SVR model. For the female subjects, the best score is 42.1\,mm with the NN model.

As described in Section \ref{sec:hme_res}, we further evaluate the capabilities of a \textbf{consistent shape without available GT AMs}. The naive approach is to use the median of the estimated inconsistent $\beta$ parameters (Tab. \ref{tab:ik}, column \emph{median}). Another approach is to use the meshes created by IK applied to the UU results, compute the AMs with B2A, calculate the median AMs and convert them to $\beta$ parameters via the A2B models. Results are displayed in Table \ref{tab:ik}, row three. For ASPset, using fixed body shape parameters from A2B models based on the measurements from UU results achieves a slightly better score than the results with inconsistent body shapes. For fit3D, the MPJPE increases by 0.9\,mm, but the A2B model results are a slightly better alternative for consistent body shapes compared to the median $\beta$ parameters. 

Further, our approach can be used to generate pseudo GT meshes for datasets with only 3D keypoint annotations. We can use these pseudo GT meshes to fine-tune HME models and increase their performance regarding the estimated keypoints for the specific dataset. However, these results are still worse than the results of our approach. We present the results in the supplementary.

%% file: sec/6_conclusion.tex
\section{Conclusion}
We address the problem of inconsistent estimated basic body shapes of humans in videos. We analyze the GT data of 3D pose and mesh datasets and find inconsistencies in their annotations. Then, we propose a family of learned \emph{A2B} models to convert 36 anthropometric measurements to SMPL-X $\beta$ parameters. This can be used to measure a human once (as it is established practice for athletes, our main focus) and use the resulting shape of the A2B model for all evaluations. With this strategy, the body shape is accurate and consistent per person. 
Evaluations show that using IK on the results of a SOTA 3D HPE model to estimate the mesh pose combined with our A2B model's shape parameters leads to superior and consistent results compared to existing HME models. Moreover, HME models also benefit from our approach. Replacing their estimated shape parameters with the A2B shape parameters leads to an improvement of their score and consistent body shapes. However, our approach based on 3D HPE still outperforms these scores.

%% file: sec/X_suppl.tex
\clearpage
\setcounter{page}{1}
\maketitlesupplementary

\section{Anthropometric Measurements}

\begin{figure}[b]
  \centering
    \includegraphics[width=0.9\linewidth]{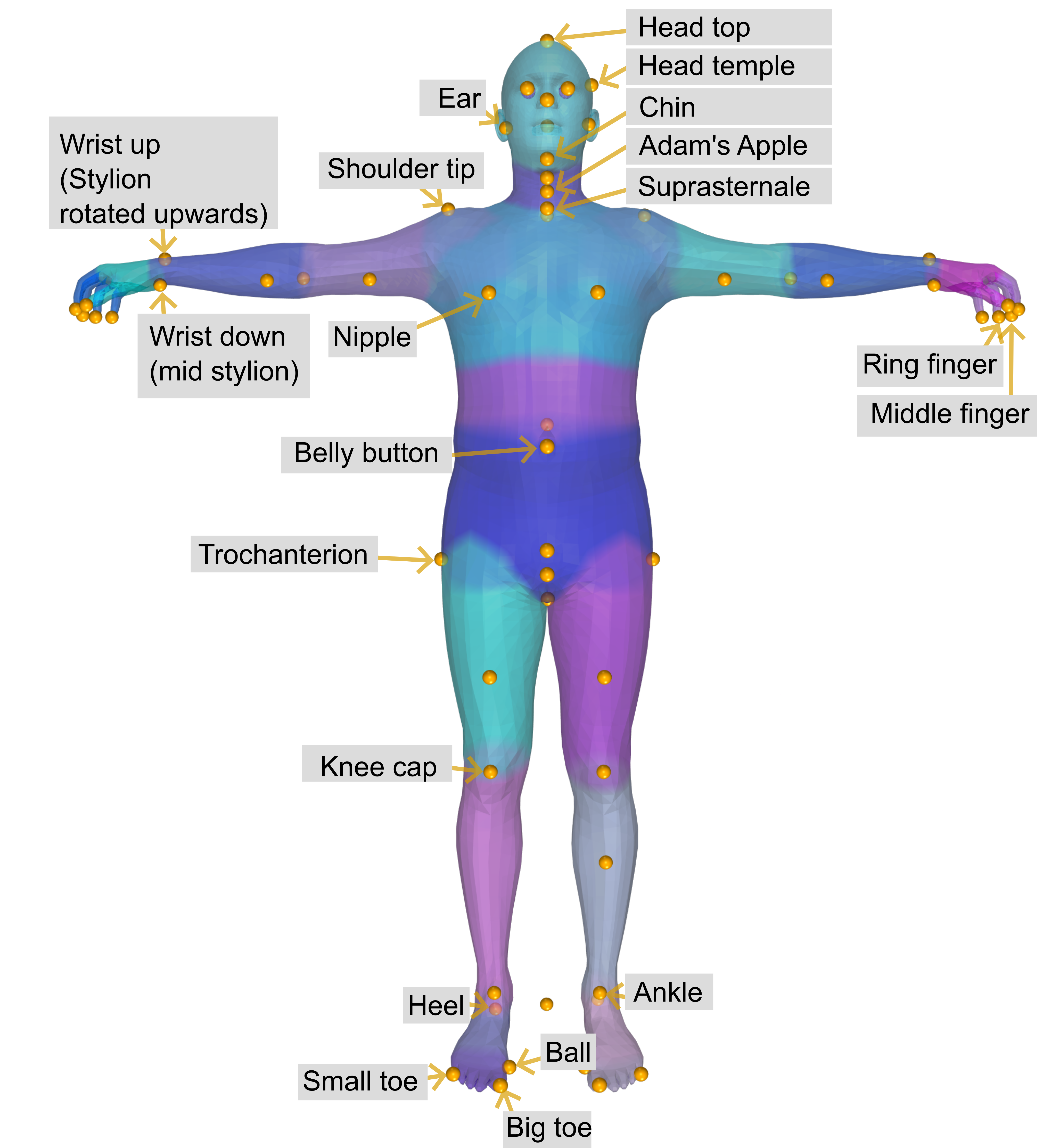}
   \caption{Visualization of the used landmarks with a standard T-pose SMPL-X mesh in front view.}
   \label{fig:landmarks_f}
\end{figure}

\begin{figure}[tb]
  \centering
    \includegraphics[width=0.35\linewidth]{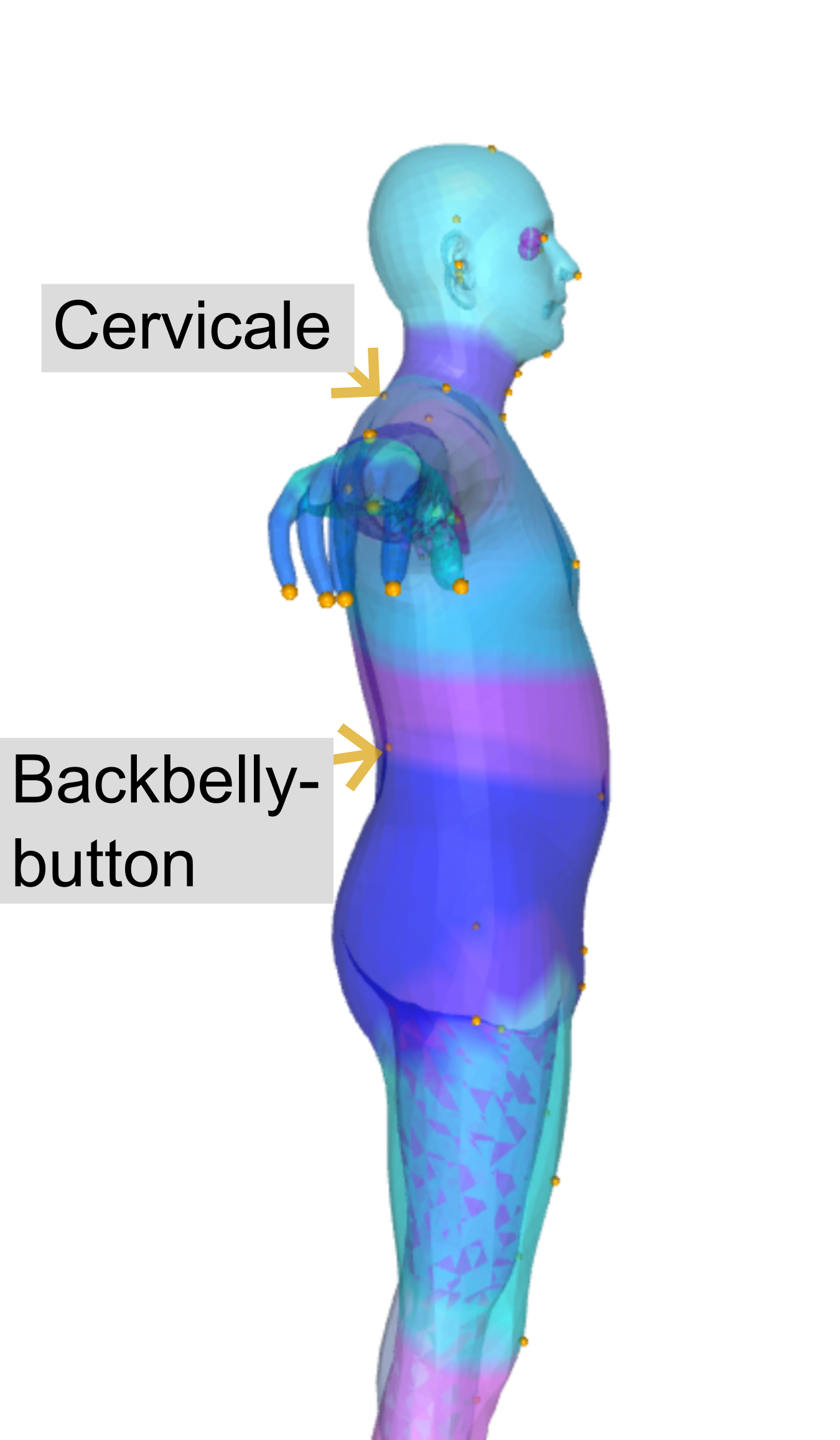}
   \caption{Visualization of a subset of the used landmarks in side view.}
   \label{fig:landmarks_s}
\vspace{-0.5cm}
\end{figure}

\begin{figure}[hb]
  \centering
    \includegraphics[width=0.5\linewidth]{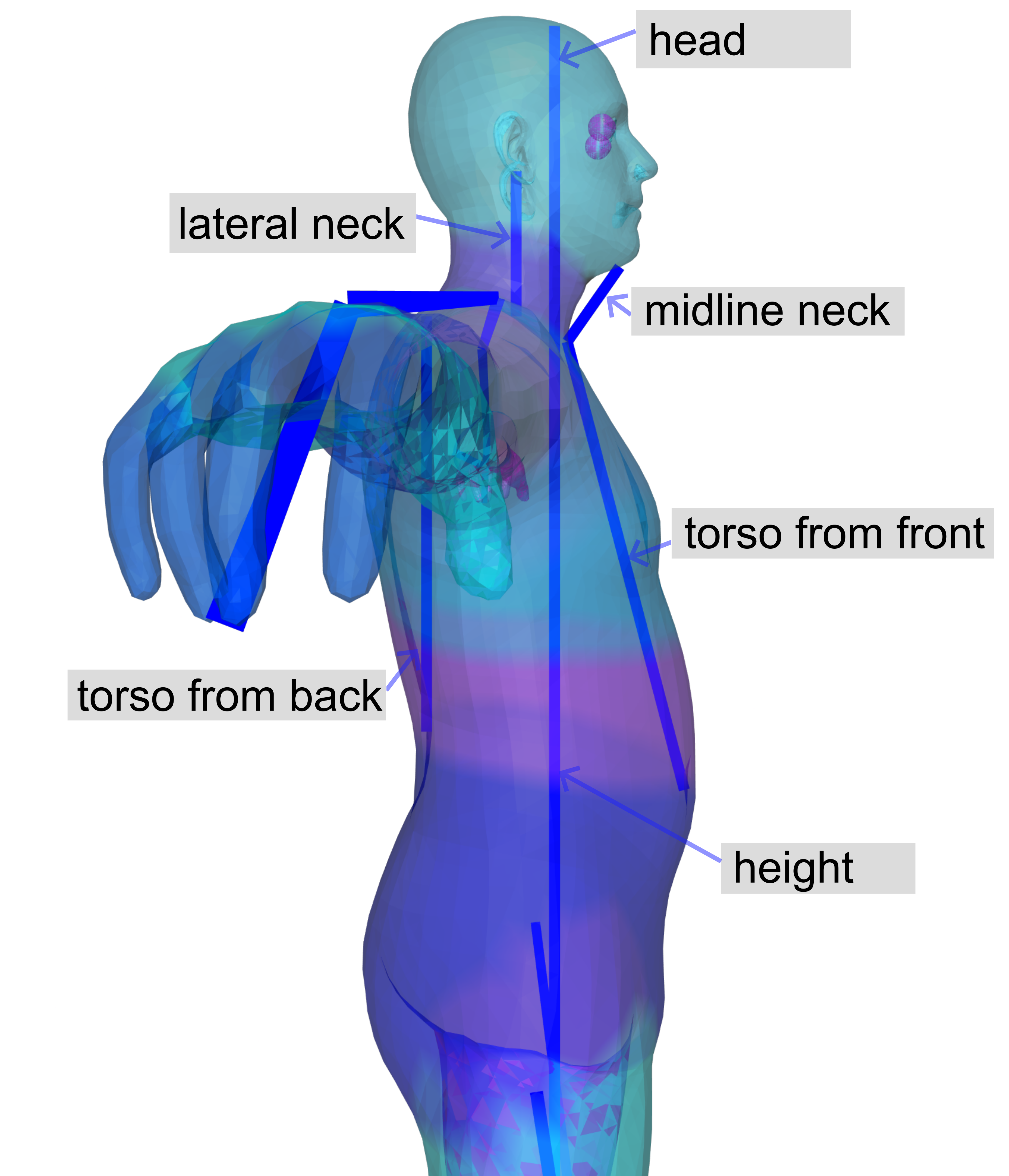}
   \caption{Visualization of used lengths with a standard T-pose SMPL-X mesh in side view.}
   \label{fig:lengths_s}
   \vspace{-0.5cm}
\end{figure}

\begin{figure}[tb]
  \centering
    \includegraphics[width=0.95\linewidth]{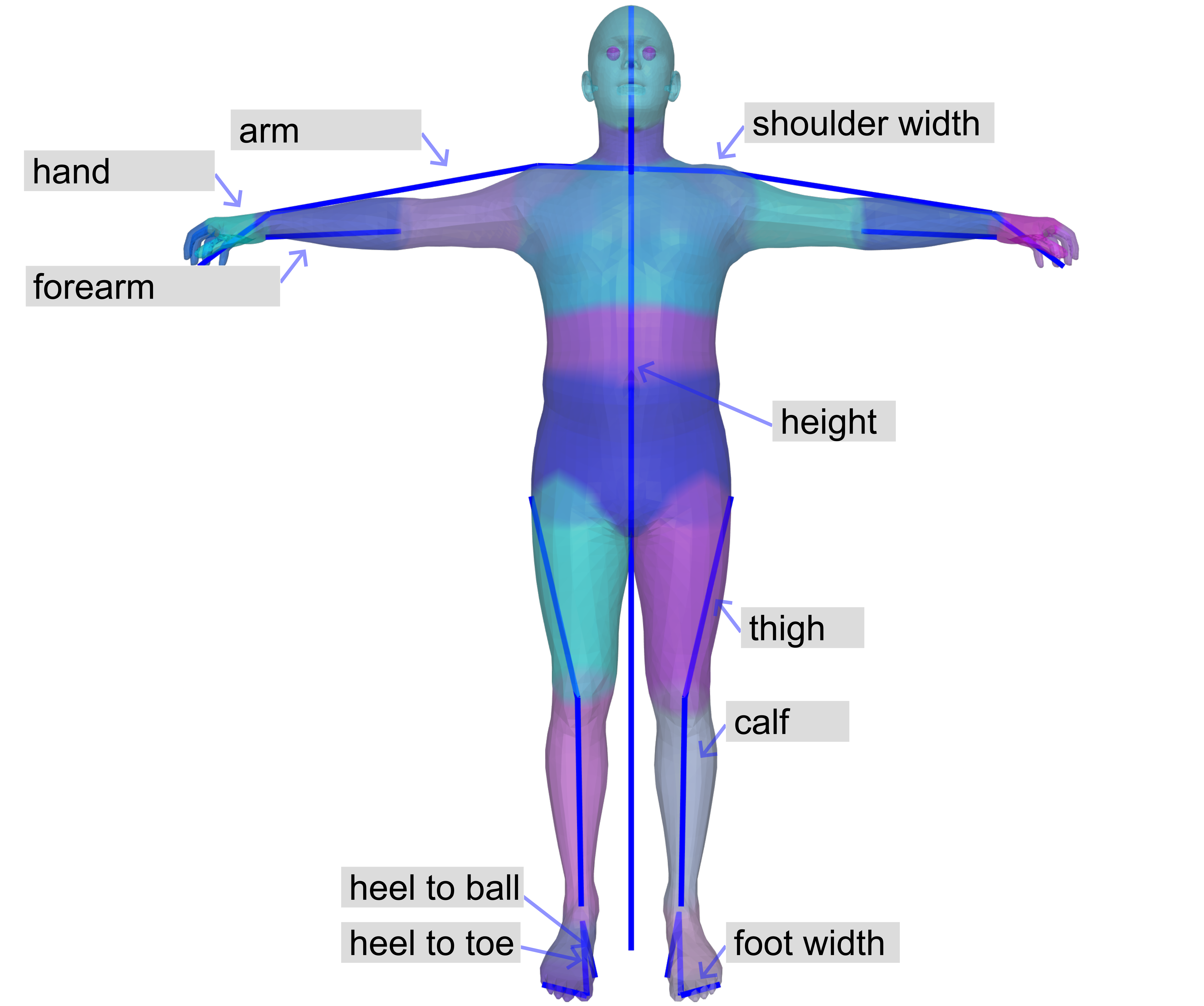}
   \caption{Visualization of used lengths with a standard T-pose SMPL-X mesh in front view.}
   \label{fig:lengths_f}
   \vspace{-0.5cm}
\end{figure}

\begin{figure}[tb]
  \centering
    \includegraphics[width=0.8\linewidth]{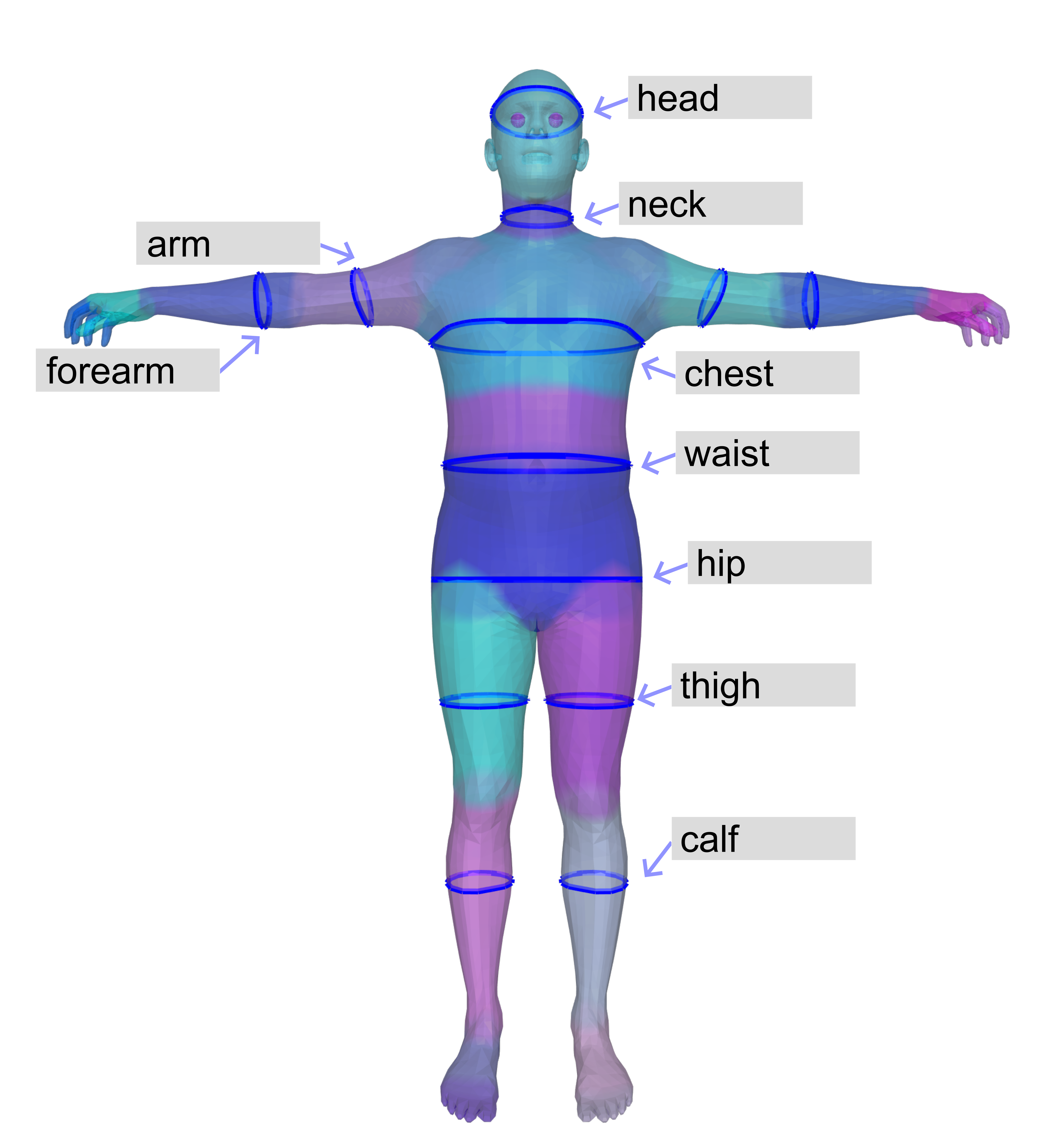}
   \caption{Visualization of used circumferences with a standard T-pose SMPL-X mesh in front view.}
   \label{fig:circumferences}
\end{figure}

The selection of the anthropometric measurements is mainly adopted from AnthroNet \citep{anthronet}. In total, 36 measurements are selected, which can be divided into 23 lengths and 13 circumferences. All measurements are taken based on the standard SMPL-X T-pose. The reference landmarks are chosen by matching the vertices on the default mesh with the landmarks defined by the anthropometric survey of the U.S. army personnel \citep{usarmy}. A visualization of the landmarks can be found in Figure \ref{fig:landmarks_f} and \ref{fig:landmarks_s}. The lengths are calculated by computing the Euclidean distance between two landmarks or the difference along the coordinate axis pointing upwards for certain heights.
The lenghts are visualized in Figure \ref{fig:lengths_s} and \ref{fig:lengths_f}. Table \ref{tab:lengths} lists the enclosing landmarks for each length. To measure the circumferences, we adopt the code from \citep{smplanthro}. For each measurement, a plane is created, the intersection between the mesh and the plane are extracted and the convex hull of the result is calculated. 
During this process, the mesh is restricted to the body part to be measured. A visualization of the circumferences can be found in Figure \ref{fig:circumferences} and a list of the landmarks and the normal vectors spanning the plane in Table \ref{tab:circumferences}.


\begin{table}[tb]
\centering
\vspace{0.2cm}
\resizebox{\linewidth}{!}{ 
\begin{tabular}{c|c|cc}
\toprule
Idx & Circumference & Normal Vector & Position \\
\midrule
1 & Waist & Up & Belly button\\
2 & Chest & Up & Nipple \\
3 & Hip & Up & Pubic bone \\
4 & Head & Up & Head temple \\
5 & Neck & Spine to head & Adam's apple \\
6/7 & Upper Arm & Shoulder to elbow & Center of the bicep \\
8/9 & Forearm & Elbow to wrist & Widest point of the forearm \\
10/11 & Thigh  & Up & Center of the thigh \\
12/13 & Calf & Up & Widest point of the calf \\
\bottomrule
\end{tabular}
}
\caption{Definitions of circumferences by landmarks and the normal vector spanning the plane.} \label{tab:circumferences}
\end{table}

\section{3D Human Shape Ground Truth Analysis}

We further analyze the GT shape consistency for the common datasets Human3.6M \citep{h36m} and MPI-INF-3DHP \citep{mpi_inf_3d}. We find that for Human3.6M, the bone lengths derived from the 3D annotations are fixed, but not for MPI-INF-3DHP. Therefore, we do not report the deviations of 3D joint annotations for Human3.6M, since there are none. We further evaluate the SMPL-X annotations for both datasets provided by NeuralAnnot \citep{neuralannot} which are used by HME models as GT for training. See Tables \ref{tab:mpi}, \ref{tab:h36m} for details.
\begin{table}[htb]
\centering
\resizebox{\linewidth}{!}{ 
\begin{tabular}{c|ccc||c|ccc}
\toprule
\multicolumn{4}{c||}{3D joint annotations} & \multicolumn{4}{c}{SMPL-X annotations}\\
Measure & $\sigma$ & r. $\sigma$& r. range & Measure & $\sigma$ & r. $\sigma$ & r. range\\
\midrule
head & 0.19 & 1.03\% & 2.08\%              	&head & 0.21 & 0.75\% & 4.87\%\\
hip width & 0.22 & 0.89\% & 1.80\% 			&hip circ. & 1.16 & 1.16\% & 9.13 \%\\
forearm & 0.21 & 0.87\% & 1.77\% 			&forearm & 0.45 & 1.80\% & 9.75\%\\
upper arm & 0.29 & 0.90\% & 1.82\% 		&arm & 0.83 & 1.59\% & 8.19\%\\
lower leg & 0.60& 1.49\% & 3.06\%			&lower leg & 1.05 & 2.56\% & 11.54\%\\
thigh & 3.83 & 7.91 \% & 41.90\%				&thigh & 0.77 & 2.02\% & 9.47\%\\
 & & &     													&height & 2.76 & 1.56\% & 8.24\% \\ 
  & & &     													&$\beta$ param. & 0.18 \\
  \bottomrule
\end{tabular}
}
\caption{GT data analysis for MPI-INF-3DHP \cite{mpi_inf_3d}. Bone length analysis based on the 3D joint locations (left) and on SMPL-X annotations by NeuralAnnot (right).  Standard deviation $\sigma$, relative standard deviation $\frac{\sigma}{\textit{avg}}$ and relative range $\frac{\max - \min}{\textit{avg}}$ of anthropometric measurements are reported. Standard deviations are given in cm, despite for the $\beta$ parameters. The values are averaged between left and right body parts and between all persons in of each dataset. The $\beta$ parameter standard deviation is averaged over all $\beta$ parameters. } \label{tab:mpi}
\vspace{-0.5cm}
\end{table}

\begin{table}[htb]
\centering
\resizebox{0.55\linewidth}{!}{ 
\begin{tabular}{c|ccc}
\toprule
 \multicolumn{4}{c}{SMPL-X annotations}\\
Measure & $\sigma$ & r. $\sigma$ & r. range\\
\midrule
head & 0.41 & 1.51\% & 10.28\%\\
hip circ. & 1.24 & 1.19\% & 8.90\%\\
forearm & 0.83 & 3.30\% & 27.93\%\\
arm & 0.77 & 2.58\% & 22.88\%\\
lower leg & 0.43 & 1.18\% & 12.20\%\\
thigh & 0.66 & 1.27\% & 9.43\%\\
height & 3.40 & 2.06\% & 15.66\% \\ 
$\beta$ param. & 0.20 \\
  \bottomrule
\end{tabular}
}
\caption{GT data analysis for Human3.6M \cite{h36m}: Analysis of SMPL-X annotations by NeuralAnnot.  Standard deviation $\sigma$, relative standard deviation $\frac{\sigma}{\textit{avg}}$ and relative range $\frac{\max - \min}{\textit{avg}}$ of anthropometric measurements  are reported. Standard deviations are given in cm, despite for the $\beta$ parameters. The values are averaged between left and right body parts and between all persons in of each dataset. The $\beta$ parameter standard deviation is averaged over all $\beta$ parameters. } \label{tab:h36m}
\vspace{-0.5cm}
\end{table}

\begin{table*}[htb]
\centering
\resizebox{0.73\linewidth}{!}{ 
\begin{tabular}{c|c|cc}
\toprule
Idx & Length Name & From & To \\
\midrule
1 & Shoulder width & Left shoulder tip (left acromion) & Right shoulder tip \\
2 & Back torso height & Cervicale & Back belly button \\
3 & Front torso height & Suprasternale (top of the breastbone) & Belly button \\
4 & Head & Head top & Cervicale \\
5 & Midline neck & Chin & Suprasternale \\
6 & Lateral neck & Center between the ears & Cervicale \\
7 & Height & Head top & Center between heels \\
8/9 & Hand right/left & Center between middle and ring finger & Stylion rotated downwards \\
10/11 & Arm right/left & Acromion & Wrist \\
12/13 & Forearm right/left & Elbow & Stylion rotated downwards\\
14/15 & Thigh right/left & Outer point at the femur (Trochanterion) & Knee cap\\
16/17 & Calf right/left & Knee cap & Ankle \\
18/19 & Foot width right/left & Small toe & Big toe\\
20/21 & Heel to ball right/left & Heel & Ball \\
22/23 & Heel to toe right/left & Heel & Big toe \\
\bottomrule
\end{tabular}
}
\caption{Definitions of lengths by their two enclosing landmarks.} \label{tab:lengths}
\end{table*}

\section{Evaluating A2B Models}

\begin{figure}[htb]
  \centering
    \includegraphics[width=\linewidth]{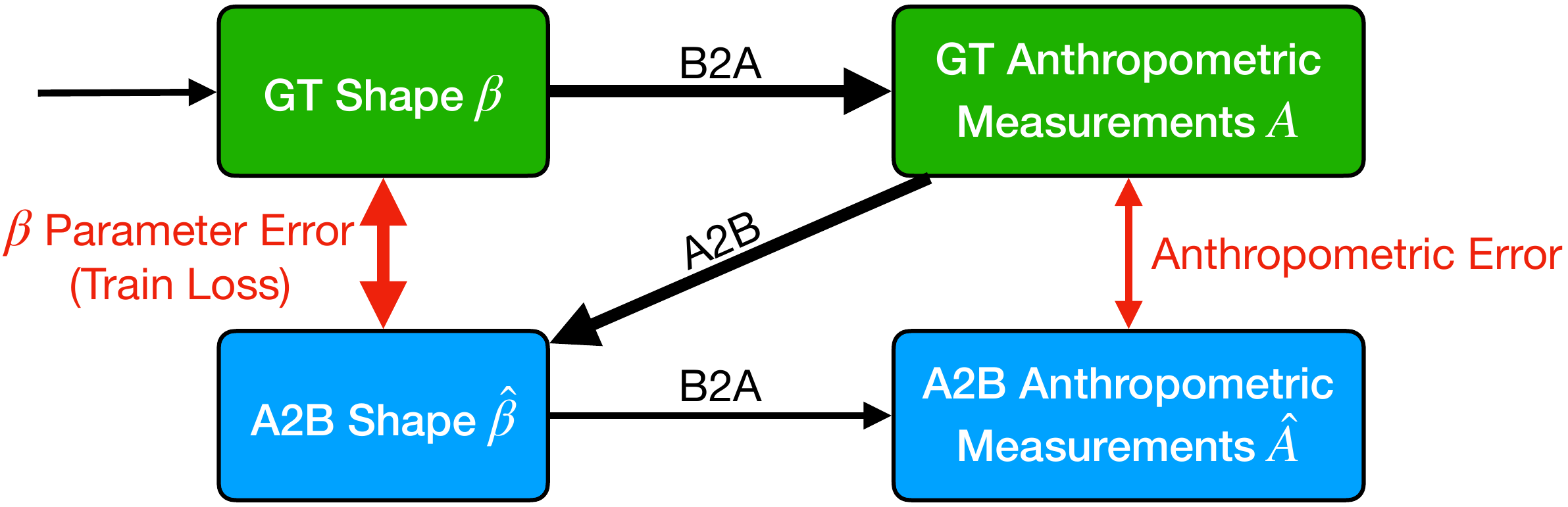}
   \caption{Visualization of the A2B evaluation and training procedures. The training part is highlighted with thicker arrows. During training, the $\beta$ parameter error is used. For evaluations, the $\beta$ parameter error and the anthropometric error are calculated.}
   \label{fig:a2b_eval}
\end{figure}

We measure two types of errors to evaluate the performance of our A2B models. The first type ($\beta$ error) shows the error if we take the GT $\beta$ parameters, derive anthropometric measurements (B2A), input them into the A2B models and evaluate the MSE of the predicted $\beta$ parameters. The second type ($A$ error) calculates B2A from the predicted $\beta$ parameters and evaluates the mean difference between the GT and predicted anthropometric measurements (all 36) in mm. These evaluations are a kind of cycle consistency evaluation for A2B and B2A. Figure \ref{fig:a2b_eval} provides a visualization of the evaluation scheme. The part that is also included in the training is highlighted with thicker arrows. The anthropometric error is only used during evaluation. 

In the main paper, we test our A2B models on the AGORA \cite{agora} dataset and randomly sampled body shapes. Since AGORA is a synthetic dataset, it might not reflect the real world. The same holds for randomly sampled body shapes. Therefore, we additionally test our best A2B models on the real-world SSP-3D dataset \cite{ssp3d} which consists of diverse body shapes. We display the results in Table \ref{tab:ssp3d}.

\begin{table}[htb]
\centering
\resizebox{0.8\linewidth}{!}{ 
\begin{tabular}{c|ccc|ccc}
\toprule
& \multicolumn{3}{c|}{$\diameter$ error of $\beta$ [$10^{-2}$]} & \multicolumn{3}{c}{$\diameter$ error of $A$ [mm]}\\
    &  m & f & n & m & f & n\\
\midrule
NN  & 1.73 & 0.97 & 2.74  &	0.634 & 0.803 & 0.968	\\
SVR & 0.13 & 0.0039 & 0.066	& 0.167 & 0.114 & 0.182	\\
    \bottomrule
\end{tabular}
}
\caption{Results of our A2B models on the SSP-3D dataset using n(eutral), m(ale) and f(emale) meshes.}
\label{tab:ssp3d}

\end{table}
All A2B models accurately estimate the diverse real-world body shapes with low error.

\section{Keypoint Selection for fit3D}

We use the fit3D \cite{fit3d} dataset for our evaluations, since this is the only sports dataset with public SMPL-X annotations. We evaluate on the SMPL-X joints, since these are trivial to obtain from SMPL-X meshes and there is no regressor available for the fit3D annotated 3D joints. SMPL-X has 144 defined joints. Since our focus is mainly on the body and not on the hands and face, we remove most of these joints. In the end, we select a subset of 37 SMPL-X joints: pelvis, left hip, right hip, spine1, left knee, right knee, spine2, left ankle, right ankle, spine3, left foot, right foot, neck, left collar, right collar, head, left shoulder, right shoulder, left elbow, right elbow, left wrist, right wrist, left index, left thumb, right index, right thumb, left big toe, left small toe, left heel, right big toe, right small toe, right heel, right eye, left eye, right ear, left ear, nose.

\section{Generation of Pseudo GT Anthropometric Measurements}

As we do not have access to the athletes of ASPset and fit3d to obtain real anthropometric measurements, we need an alternative to simulate this process. For ASPset, as a first step, we run IK on the GT 3D joint locations. From the generated meshes, we obtain the necessary anthropometric parameters with B2A. Then, we use the median values of these measurements as the GT anthropometric values. We call these parameters \emph{pseudo GT} throughout this paper, since this is not directly the GT, but obtained from IK executed on the GT 3D joint locations and the B2A computation from the created meshes. These parameters are used in this paper to generate the pseudo GT $\beta$ parameters by A2B prediction. 

\begin{table*}[tb]
  \centering
  \resizebox{0.75\linewidth}{!}{ 
  \begin{tabular}{cc|ccccc|c}
  \toprule
  pose & orig. & measure & NN m & SVR m & NN n & SVR n & median \\
  \midrule
  SMPLer-X  & 86.02  & SMPLer-X  & 85.89 & 85.69 & 86.02 & 85.99 & 86.04\\
  SMPLer-X FT & 79.09 & SMPLer-X FT & 78.92 & 78.88 & 79.59 & 79.37 & 79.44 \\
  SMPLer-X FT & - & GT & 65.63 & 65.84 &{64.77} & \textbf{64.76} & - \\
  SMPLer-X FT & - & SMPLer-X & 73.41 & 73.29 & 73.79 & 73.63 & 73.66 \\
  \midrule
  IK-UU &  67.54 & IK-UU & 66.92 & 66.60 & 67.28 & 67.12 & 67.16 \\
  IK-UU &  - & SMPLer-X & 63.80 & \underline{63.64} & 63.92 & 63.78 & 63.82 \\
  IK-UU & - &  SMPLer-X FT & 69.46 & 69.27 & 69.83 & 69.63 & 69.69 \\
  IK-UU & - & GT & 56.44 & 56.56 & \textbf{55.18} & 55.19 & - \\
   \bottomrule
  \end{tabular}
  }
  \caption{MPJPE results in mm for the test split of ASPset. Results are given for different methods and replaced $beta$ parameters with A2B results (columns NN/SVR) or the median of the original $\beta$ parameters from the model noted in the \emph{measure} column. SMPLer-X FT stands for the best fine-tuned variant of SMPLer-X (fine-tuned with the meshes obtained from IK executed on the GT 3D joints). The \emph{orig} column contains the results without replaced $\beta$ parameters. We highlight the best result for each model and the best option for the combination of IK-UU pose and SMPLer-X $\beta$ parameters, since this combination outperforms the original IK-UU result, too.\label{tab:finetune}}
  \end{table*}
We do not have access to the athletes of the fit3D dataset either. Therefore, we need some kind of GT data to mimic measurements. Obviously, there is no GT available for the official test set evaluation on the evaluation server. We therefore split the official training dataset into a training, validation, and test set for our evaluations. We perform a leave-one-out cross validation, therefore all eight athletes from the official training dataset are used in our evaluation.
With this selection, we have real GT shape parameters available. We do not use these directly, since this would skip the measuring process that is needed in real applications. Further, the GT data is not consistent (see Section 3 in the main paper). Therefore, we apply B2A and use the median measurements over time in order to simulate the measuring process and obtain a single set of anthropometric measurements per person. In real applications, this step is omitted because the anthropometric parameters can be measured directly from the athletes before starting the recording. 

We consider this strategy as a valid method for evaluations, since our main goal is to improve the HME performance as much as possible with only marginal overhead. Our main focus is sports, which contains extreme poses that let existing HME models fail, sometimes even to detect a human at all. Examples can be found in the supplementary videos. As professional athletes are measured anyway, the additional effort for the measurements is negligible in this context.

\section{Inverse Kinematics}

We use the inverse kinematics approach with a VPoser extension, as proposed in the code by \citep{smplx}, to fit SMPLX meshes to given 3D keypoints. VPoser is a learned prior for human poses, since the raw SMPL-X model definition allows impossible poses for humans. VPoser learned plausible poses from the large AMASS \citep{amass} dataset and helps IK to generate only plausible poses. IK learns the best SMPL-X parameters ($\beta$ and $\theta$) that fit the mesh to the given 3D joint locations by minimizing the error between the given joint locations and the regressed joint locations from the mesh.
IK is an iterative algorithm and adjusts the pose and the shape parameters with a gradient descent minimization approach in each step. Besides the already described joint error, IK further penalizes abnormal poses with a VPoser error and extreme body shapes with a $\beta$ parameter error. Therefore, the total loss for IK can be described as:
\begin{equation}
\mathcal{L}_{\mathit{IK}} = \lambda_1 \mathcal{L}_{\mathit{joint}} + \lambda_2 \mathcal{L}_{\mathit{VPoser}} + \lambda_3 \mathcal{L}_{\beta},
\end{equation}
whereby $\mathcal{L}_{\mathit{joint}}$ is the summarized Squared Error of the estimated keypoints, $\mathcal{L}_{\mathit{VPoser}}$ and $\mathcal{L}_{\beta}$ are the sums of the squared values of the VPoser and $\beta$ parameters, respectively. This makes sense since the VPoser and $\beta$ parameter distributions are centered around zero. We set the weighting factors $\lambda_1 = 10$, $\lambda_2 = 0.0007$, and $\lambda_3 = 0.01$ in our experiments. We use relatively low values for $\lambda_2$ and $\lambda_3$, since sports datasets incorporate extreme poses and our main interest is to achieve the most perfect pose.

We execute IK per frame, which results in a slight jitter in between the frames, but leads to more accurate joint positions. Since IK needs multiple iterations to adjust the standard T-pose parameters to achieve a pose that is roughly close to the desired UU pose, we speed up the process by initializing the pose and shape parameters with the result from the previous frame if available. This also enhances the final result slightly. We acknowledge that IK is relatively slow regarding the runtime, but our main focus is the precision. For sport analysis, which is our focus, the runtime is not critical, but a very precise result is crucial.

\section{Fine-tuning HME Models with Pseudo GT Meshes}

Fine-tuning existing HME models on pure 3D joints datasets is not possible, since they need mesh annotations for training. However, with IK, we can generate pseudo GT meshes. We exemplary test a fine-tuning of SMPLer-X on ASPset with this approach. Experiments show that using their fine-tuning script with 1.6M iterations leads to worse results than the results without fine-tuning. Therefore, we reduce the number of iterations with early stopping and achieve better results with fine-tuning only for 32K iterations. 

The results shown in Table \ref{tab:finetune} prove that fine-tuning on IK generated meshes can lead to a significant improvement of the scores. Replacing the $\beta$ parameters of the fine-tuned results with the A2B $\beta$ parameters boosts the performance even more. These are the best results achieved with any existing HME model throughout this study. 

Moreover, we experiment with using the SMPLer-X body shape parameters combined with the poses estimated by IK applied to the UU results (see last two rows of Table \ref{tab:finetune}). Using the $\beta$ parameters from SMPLer-X leads to a slightly better result than the original 3D joint based result (without IK). This evaluation shows that 3D HPE models are better in precisely locating the joints of humans than HME models, but HME models are better in estimating the shape of humans. We also try to use the $\beta$ parameters of the fine-tuned variant together with the UU IK poses like before. However, this resulted in a performance drop compared to the body shape parameters from the original SMPLer-X without fine-tuning. These experiments show that fine-tuning HME models on pseudo ground truth leads to a better performance regarding the keypoints, but the estimated $\beta$ parameters have worse quality. This can further be proven by replacing the $\beta$ parameters from the fine-tuned SMPLer-X variant with the $\beta$ parameters from the not fine-tuned model, which results in a performance gain of over 5\,mm compared to the original results from the fine-tuned version (rows 2 and 4 in Tab. \ref{tab:finetune}). However, our method using the UU IK poses and the A2B body shape parameters with GT anthropometric measurements achieves the overall best results.

We provide a comprehensive summary and visualization of all results on the ASPset dataset in Section \ref{sec:summary_results}. This includes results of existing HME models, results of our approach, and the fine-tuning results.

\section{Summary of the Results}\label{sec:summary_results}

\begin{figure}[htb]
  \centering 
    \includegraphics[width=\linewidth]{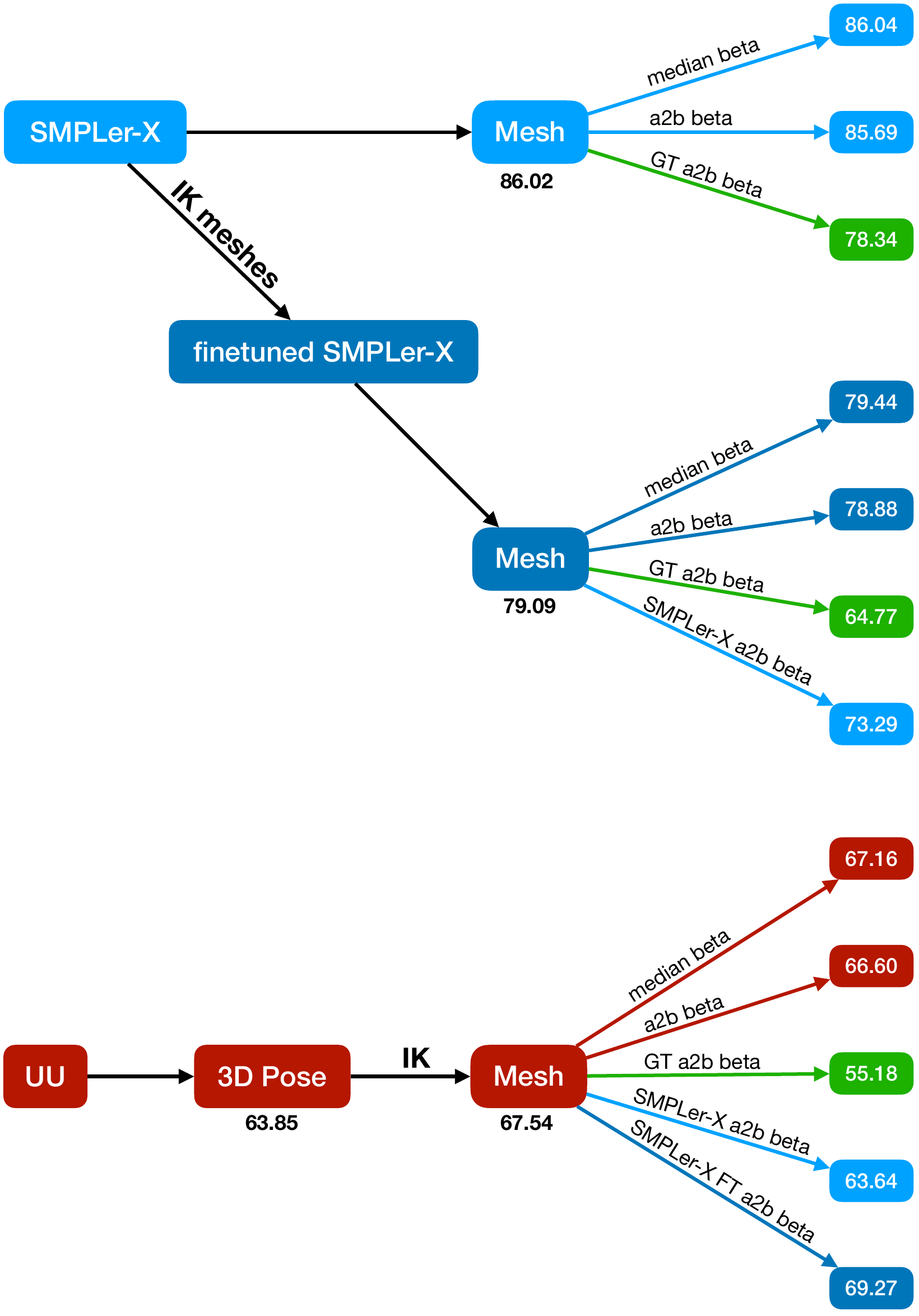}
  \hfill
   \caption{Overview of the main results for the ASPset dataset. All results are MPJPE results in mm. Results below \emph{mesh} boxes show the result with the original $\beta$ parameters. All results after arrows to the right are results with replaced $\beta$ parameters. The type of the $\beta$ parameters is noted on the arrow and is color-coded.}
   \label{fig:overview}
\end{figure}
\begin{figure*}[htb]
  \centering
  \begin{subfigure}[b]{0.49\linewidth}
    \includegraphics[height=0.99\linewidth]{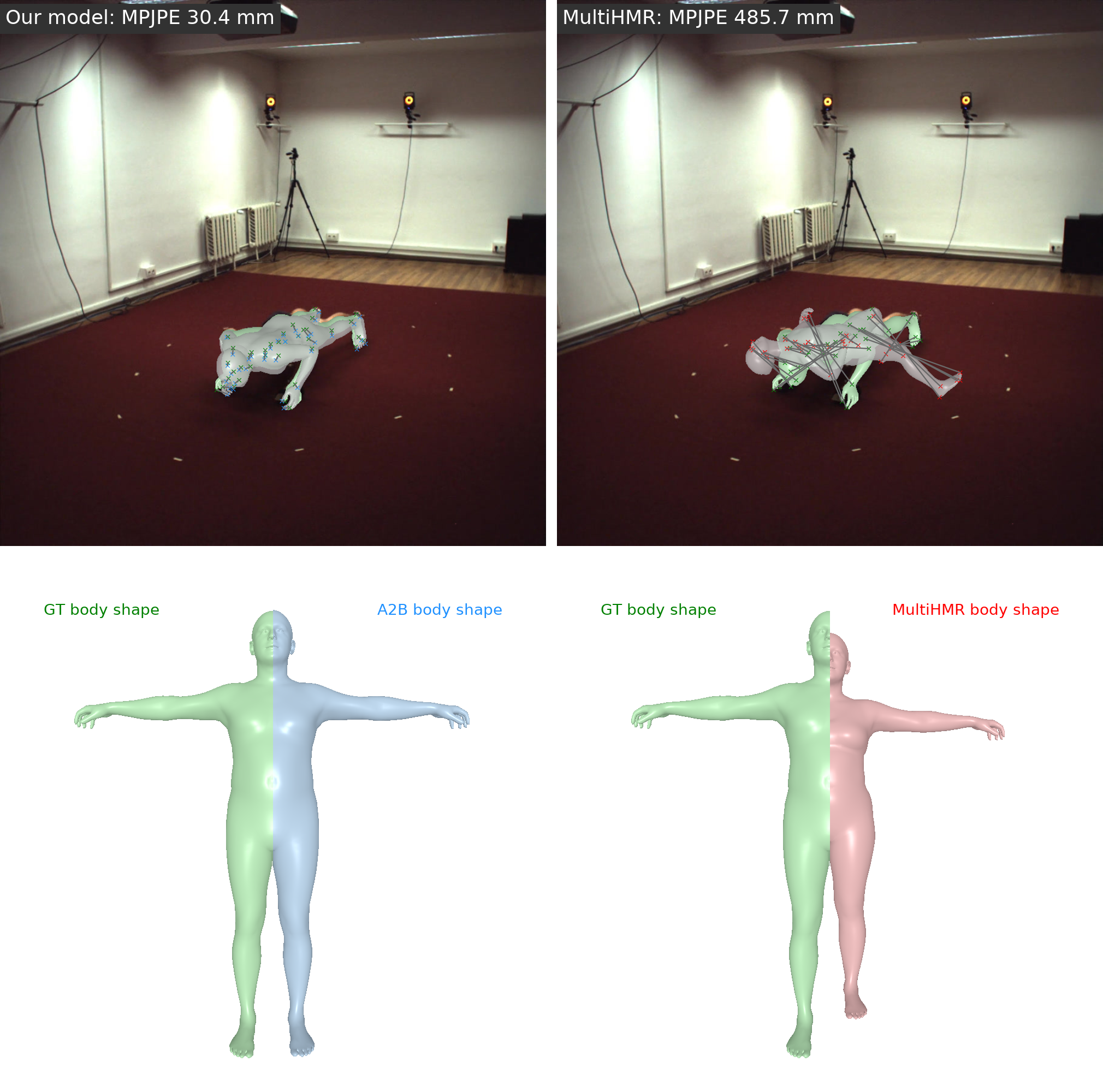}
  \end{subfigure}
  \hfill
  \begin{subfigure}[b]{0.49\linewidth}
    \includegraphics[height=0.99\linewidth]{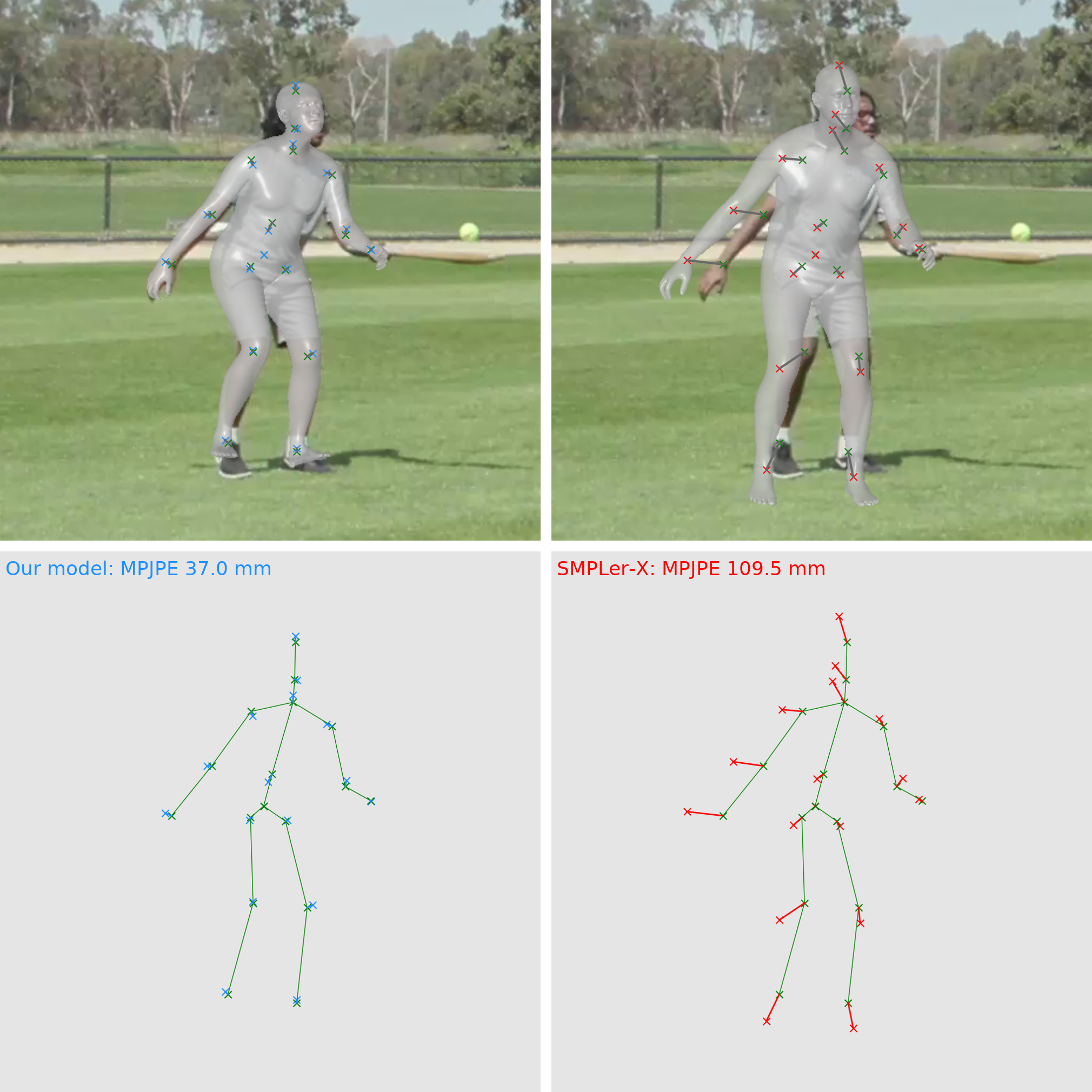}
  \end{subfigure}
  \hfill
   \caption{Example frames from our supplementary videos. It shows qualitative results of our approach compared to MultiHMR for fit3d (left) and qualitative results of SMPLer-X and our approach for example frames from ASPset (right). In fit3d visualizations, we display the \textcolor{ForestGreen}{GT meshes in green} and the estimated meshes in gray. The GT joints are also displayed in green while the estimated joints from our model are visualized in \textcolor{blue}{blue}. The \textcolor{red}{MultiHMR joints are shown in red}. Corresponding joints are connected. We display the exact MPJPE values in the top left of each frame. Recall that the visualization is in 2D, but the evaluation is in 3D. Therefore, sometimes the MPJPE values may seem odd. In the lower part, we show the estimated body shapes in T-pose. The \textcolor{ForestGreen}{GT body shape is shown in green} and the estimated body shape from \textcolor{blue}{our model in blue}. The \textcolor{red}{MultiHMR body shape is shown in red}. For ASPset visualizations, we display the estimated meshes and the GT and estimated joints. \textcolor{ForestGreen}{GT joints are shown in green}, estimated joints from \textcolor{blue}{our model in blue.} and the \textcolor{red}{SMPLer-X joints in red}. Corresponding joints are connected. In the lower part, we show the GT and estimated joints in the same way, but without the mesh and image to reduce distraction. We further display the MPJPE values.}
   \label{fig:supp_vis}
\end{figure*}
We execute a multitude of experiments with different combinations of pose and shape parameters. Figure \ref{fig:overview} summarizes the results with their pose and shape origins for ASPset. In general, the poses estimated by IK based on the UU results (red branch in Fig. \ref{fig:overview}) are more precise than the poses estimated by SMPLer-X (light blue branch in Fig. \ref{fig:overview}). Further, the body shape parameters from our A2B models with GT anthropometric measurements (green boxes in Fig. \ref{fig:overview}) achieve the best results for all poses.
We provide more qualitative examples comparing SMPLer-X with this approach in the supplementary video. Without access to the GT, all models benefit slightly from A2B model results with the median anthropometric measurements from B2A of the estimated meshes by the respective model (boxes with same color for all three branches in Fig. \ref{fig:overview}). 
Moreover, SMPLer-X A2B body shape parameters perform best when analyzing body shapes without GT access (light blue boxes in Fig. \ref{fig:overview}). Fine-tuning SMPLer-X with IK created meshes (dark blue branch in Fig. \ref{fig:overview}) improves the performance of SMPLer-X, although the quality of the body shape deteriorates. This can be seen as by comparing the shapes from SMPLer-X and fine-tuned SMPLer-X (dark blue and light blue boxes in Fig. \ref{fig:overview}) with fine-tuned and IK poses.

Since fit3D is a larger dataset, fine-tuning UU works better, which further leads to better IK meshes with an MPJPE of 37.02\,mm. Enforcing consistent meshes with GT or IK A2B shape parameters decreases the performance slightly in this case. However, A2B shape parameters achieve slightly better scores than median values. This also holds for OSX and Multi-HMR. Overall, the approach with UU, IK, and A2B body shape parameters achieves an over 33\,mm lower MPJPE than any HME model. The same also holds for the MVE, which can be improved by over 30\,mm with our approach. The scores can be found in the main paper.

We provide two videos in the supplementary material that show qualitative results for ASPset and fit3D. Figure \ref{fig:supp_vis} shows one example visualization for both datasets. We include the GT and predicted meshes in the fit3d visualization and display the GT and estimated body shapes in T-pose right next to each other. For ASPset, we visualize the estimated meshes and the GT and estimated joints, since we do not have GT meshes here.

%% file: main.bbl
\begin{thebibliography}{38}
\providecommand{\natexlab}[1]{#1}
\providecommand{\url}[1]{\texttt{#1}}
\expandafter\ifx\csname urlstyle\endcsname\relax
  \providecommand{\doi}[1]{doi: #1}\else
  \providecommand{\doi}{doi: \begingroup \urlstyle{rm}\Url}\fi

\bibitem[Baradel et~al.(2024)Baradel, Armando, Galaaoui, Br{\'e}gier,
  Weinzaepfel, Rogez, and Lucas]{multihmr}
Fabien Baradel, Matthieu Armando, Salma Galaaoui, Romain Br{\'e}gier, Philippe
  Weinzaepfel, Gr{\'e}gory Rogez, and Thomas Lucas.
\newblock Multi-hmr: Multi-person whole-body human mesh recovery in a single
  shot.
\newblock \emph{arXiv preprint arXiv:2402.14654}, 2024.

\bibitem[Bojanic(2023)]{smplanthro}
David Bojanic.
\newblock Smpl-anthropometry.
\newblock https://github.com/DavidBoja/SMPL-Anthropometry/, 2023.

\bibitem[Cai et~al.(2024)Cai, Yin, Zeng, Wei, Sun, Yanjun, Pang, Mei, Zhang,
  Zhang, et~al.]{smplerx}
Zhongang Cai, Wanqi Yin, Ailing Zeng, Chen Wei, Qingping Sun, Wang Yanjun,
  Hui~En Pang, Haiyi Mei, Mingyuan Zhang, Lei Zhang, et~al.
\newblock Smpler-x: Scaling up expressive human pose and shape estimation.
\newblock \emph{Advances in Neural Information Processing Systems}, 36, 2024.

\bibitem[Cha et~al.(2022)Cha, Saqlain, Kim, Shin, and Baek]{3dhmeik}
Junuk Cha, Muhammad Saqlain, GeonU Kim, Mingyu Shin, and Seungryul Baek.
\newblock Multi-person 3d pose and shape estimation via inverse kinematics and
  refinement.
\newblock In \emph{European Conference on Computer Vision}, pages 660--677.
  Springer, 2022.

\bibitem[Chen et~al.(2021)Chen, Pang, Yang, Ma, Xu, and Yu]{sportscap}
Xin Chen, Anqi Pang, Wei Yang, Yuexin Ma, Lan Xu, and Jingyi Yu.
\newblock Sportscap: Monocular 3d human motion capture and fine-grained
  understanding in challenging sports videos.
\newblock \emph{International Journal of Computer Vision}, 129:\penalty0
  2846--2864, 2021.

\bibitem[Choi et~al.(2022)Choi, Moon, Park, and Lee]{3DCrowdNet}
Hongsuk Choi, Gyeongsik Moon, JoonKyu Park, and Kyoung~Mu Lee.
\newblock Learning to estimate robust 3d human mesh from in-the-wild crowded
  scenes.
\newblock In \emph{Proceedings of the IEEE/CVF Conference on Computer Vision
  and Pattern Recognition}, pages 1475--1484, 2022.

\bibitem[Choutas et~al.(2020)Choutas, Pavlakos, Bolkart, Tzionas, and
  Black]{expose}
Vasileios Choutas, Georgios Pavlakos, Timo Bolkart, Dimitrios Tzionas, and
  Michael~J Black.
\newblock Monocular expressive body regression through body-driven attention.
\newblock In \emph{Computer Vision--ECCV 2020: 16th European Conference,
  Glasgow, UK, August 23--28, 2020, Proceedings, Part X 16}, pages 20--40.
  Springer, 2020.

\bibitem[Choutas et~al.(2022)Choutas, M{\"u}ller, Huang, Tang, Tzionas, and
  Black]{shapy}
Vasileios Choutas, Lea M{\"u}ller, Chun-Hao~P Huang, Siyu Tang, Dimitrios
  Tzionas, and Michael~J Black.
\newblock Accurate 3d body shape regression using metric and semantic
  attributes.
\newblock In \emph{Proceedings of the IEEE/CVF Conference on Computer Vision
  and Pattern Recognition}, pages 2718--2728, 2022.

\bibitem[Doyon et~al.(2023)Doyon, Faure, Sanz, Daura, Cassard, and
  d’Errico]{tailor}
Luc Doyon, Thomas Faure, Montserrat Sanz, Joan Daura, Laura Cassard, and
  Francesco d’Errico.
\newblock A 39,600-year-old leather punch board from canyars, gavà, spain.
\newblock \emph{Science Advances}, 9\penalty0 (15):\penalty0 eadg0834, 2023.

\bibitem[Einfalt et~al.(2023)Einfalt, Ludwig, and Lienhart]{uu}
Moritz Einfalt, Katja Ludwig, and Rainer Lienhart.
\newblock Uplift and upsample: Efficient 3d human pose estimation with
  uplifting transformers.
\newblock In \emph{Proceedings of the IEEE/CVF Winter Conference on
  Applications of Computer Vision}, pages 2903--2913, 2023.

\bibitem[Feng et~al.(2021)Feng, Choutas, Bolkart, Tzionas, and Black]{pixie}
Yao Feng, Vasileios Choutas, Timo Bolkart, Dimitrios Tzionas, and Michael~J
  Black.
\newblock Collaborative regression of expressive bodies using moderation.
\newblock In \emph{2021 International Conference on 3D Vision (3DV)}, pages
  792--804. IEEE, 2021.

\bibitem[Fieraru et~al.(2021)Fieraru, Zanfir, Pirlea, Olaru, and
  Sminchisescu]{fit3d}
Mihai Fieraru, Mihai Zanfir, Silviu-Cristian Pirlea, Vlad Olaru, and Cristian
  Sminchisescu.
\newblock Aifit: Automatic 3d human-interpretable feedback models for fitness
  training.
\newblock In \emph{The IEEE/CVF Conference on Computer Vision and Pattern
  Recognition (CVPR)}, 2021.

\bibitem[Goel et~al.(2023)Goel, Pavlakos, Rajasegaran, Kanazawa, and
  Malik]{hmr2}
Shubham Goel, Georgios Pavlakos, Jathushan Rajasegaran, Angjoo Kanazawa, and
  Jitendra Malik.
\newblock Humans in 4d: Reconstructing and tracking humans with transformers.
\newblock In \emph{Proceedings of the IEEE/CVF International Conference on
  Computer Vision}, pages 14783--14794, 2023.

\bibitem[Gordon et~al.(2014)Gordon, Blackwell, Bradtmiller, Parham, Barrientos,
  Paquette, Corner, Carson, Venezia, Rockwell, et~al.]{usarmy}
Claire~C Gordon, Cynthia~L Blackwell, Bruce Bradtmiller, Joseph~L Parham,
  Patricia Barrientos, Stephen~P Paquette, Brian~D Corner, Jeremy~M Carson,
  Joseph~C Venezia, Belva~M Rockwell, et~al.
\newblock 2012 anthropometric survey of us army personnel: Methods and summary
  statistics.
\newblock \emph{Army Natick Soldier Research Development and Engineering Center
  MA, Tech. Rep}, 2014.

\bibitem[Ingwersen et~al.(2023)Ingwersen, Mikkelstrup, Jensen, Hannemose, and
  Dahl]{sportspose}
Christian~Keilstrup Ingwersen, Christian~M{\o}ller Mikkelstrup,
  Janus~N{\o}rtoft Jensen, Morten~Rieger Hannemose, and Anders~Bjorholm Dahl.
\newblock Sportspose-a dynamic 3d sports pose dataset.
\newblock In \emph{Proceedings of the IEEE/CVF Conference on Computer Vision
  and Pattern Recognition}, pages 5219--5228, 2023.

\bibitem[Ionescu et~al.(2014)Ionescu, Papava, Olaru, and Sminchisescu]{h36m}
Catalin Ionescu, Dragos Papava, Vlad Olaru, and Cristian Sminchisescu.
\newblock Human3.6m: Large scale datasets and predictive methods for 3d human
  sensing in natural environments.
\newblock \emph{IEEE Transactions on Pattern Analysis and Machine
  Intelligence}, 36\penalty0 (7):\penalty0 1325--1339, 2014.

\bibitem[Li et~al.(2021)Li, Xu, Chen, Bian, Yang, and Lu]{hybrik}
Jiefeng Li, Chao Xu, Zhicun Chen, Siyuan Bian, Lixin Yang, and Cewu Lu.
\newblock Hybrik: A hybrid analytical-neural inverse kinematics solution for 3d
  human pose and shape estimation.
\newblock In \emph{Proceedings of the IEEE/CVF Conference on Computer Vision
  and Pattern Recognition}, pages 3383--3393, 2021.

\bibitem[Li et~al.(2023)Li, Bian, Xu, Chen, Yang, and Lu]{hybrikx}
Jiefeng Li, Siyuan Bian, Chao Xu, Zhicun Chen, Lixin Yang, and Cewu Lu.
\newblock Hybrik-x: Hybrid analytical-neural inverse kinematics for whole-body
  mesh recovery.
\newblock \emph{arXiv preprint arXiv:2304.05690}, 2023.

\bibitem[Lin et~al.(2023)Lin, Zeng, Wang, Zhang, and Li]{osx}
Jing Lin, Ailing Zeng, Haoqian Wang, Lei Zhang, and Yu Li.
\newblock One-stage 3d whole-body mesh recovery with component aware
  transformer.
\newblock In \emph{Proceedings of the IEEE/CVF Conference on Computer Vision
  and Pattern Recognition}, pages 21159--21168, 2023.

\bibitem[Loper et~al.(2015)Loper, Mahmood, Romero, Pons-Moll, and Black]{smpl}
Matthew Loper, Naureen Mahmood, Javier Romero, Gerard Pons-Moll, and Michael~J
  Black.
\newblock Smpl: a skinned multi-person linear model.
\newblock \emph{ACM Transactions on Graphics (TOG)}, 34\penalty0 (6):\penalty0
  1--16, 2015.

\bibitem[Mahmood et~al.(2019)Mahmood, Ghorbani, Troje, Pons-Moll, and
  Black]{amass}
Naureen Mahmood, Nima Ghorbani, Nikolaus~F Troje, Gerard Pons-Moll, and
  Michael~J Black.
\newblock Amass: Archive of motion capture as surface shapes.
\newblock In \emph{Proceedings of the IEEE/CVF international conference on
  computer vision}, pages 5442--5451, 2019.

\bibitem[Mehta et~al.(2017)Mehta, Rhodin, Casas, Fua, Sotnychenko, Xu, and
  Theobalt]{mpi_inf_3d}
Dushyant Mehta, Helge Rhodin, Dan Casas, Pascal Fua, Oleksandr Sotnychenko,
  Weipeng Xu, and Christian Theobalt.
\newblock Monocular 3d human pose estimation in the wild using improved cnn
  supervision.
\newblock In \emph{2017 international conference on 3D vision (3DV)}, pages
  506--516. IEEE, 2017.

\bibitem[Moon et~al.(2022{\natexlab{a}})Moon, Choi, and Lee]{hand4whole}
Gyeongsik Moon, Hongsuk Choi, and Kyoung~Mu Lee.
\newblock Accurate 3d hand pose estimation for whole-body 3d human mesh
  estimation.
\newblock In \emph{Proceedings of the IEEE/CVF Conference on Computer Vision
  and Pattern Recognition}, pages 2308--2317, 2022{\natexlab{a}}.

\bibitem[Moon et~al.(2022{\natexlab{b}})Moon, Choi, and Lee]{neuralannot}
Gyeongsik Moon, Hongsuk Choi, and Kyoung~Mu Lee.
\newblock Neuralannot: Neural annotator for 3d human mesh training sets.
\newblock In \emph{Computer Vision and Pattern Recognition Workshop (CVPRW)},
  2022{\natexlab{b}}.

\bibitem[Nibali et~al.(2021)Nibali, Millward, He, and Morgan]{aspset}
Aiden Nibali, Joshua Millward, Zhen He, and Stuart Morgan.
\newblock Aspset: An outdoor sports pose video dataset with 3d keypoint
  annotations.
\newblock \emph{Image and Vision Computing}, 111:\penalty0 104196, 2021.

\bibitem[Patel et~al.(2021)Patel, Huang, Tesch, Hoffmann, Tripathi, and
  Black]{agora}
Priyanka Patel, Chun-Hao~P Huang, Joachim Tesch, David~T Hoffmann, Shashank
  Tripathi, and Michael~J Black.
\newblock Agora: Avatars in geography optimized for regression analysis.
\newblock In \emph{Proceedings of the IEEE/CVF Conference on Computer Vision
  and Pattern Recognition}, pages 13468--13478, 2021.

\bibitem[Pavlakos et~al.(2019)Pavlakos, Choutas, Ghorbani, Bolkart, Osman,
  Tzionas, and Black]{smplx}
Georgios Pavlakos, Vasileios Choutas, Nima Ghorbani, Timo Bolkart, Ahmed~AA
  Osman, Dimitrios Tzionas, and Michael~J Black.
\newblock Expressive body capture: 3d hands, face, and body from a single
  image.
\newblock In \emph{Proceedings of the IEEE/CVF conference on computer vision
  and pattern recognition}, pages 10975--10985, 2019.

\bibitem[Picetti et~al.(2023)Picetti, Deshpande, Leban, Shahtalebi, Patel,
  Jing, Wang, Metze~III, Sun, Laidlaw, et~al.]{anthronet}
Francesco Picetti, Shrinath Deshpande, Jonathan Leban, Soroosh Shahtalebi, Jay
  Patel, Peifeng Jing, Chunpu Wang, Charles Metze~III, Cameron Sun, Cera
  Laidlaw, et~al.
\newblock Anthronet: Conditional generation of humans via anthropometrics.
\newblock \emph{arXiv preprint arXiv:2309.03812}, 2023.

\bibitem[Qiu et~al.(2022)Qiu, Yang, Wang, and Fu]{GRM3D}
Zhongwei Qiu, Qiansheng Yang, Jian Wang, and Dongmei Fu.
\newblock Dynamic graph reasoning for multi-person 3d pose estimation.
\newblock In \emph{Proceedings of the 30th ACM International Conference on
  Multimedia}, pages 3521--3529, 2022.

\bibitem[Qiu et~al.(2023)Qiu, Yang, Wang, Feng, Han, Ding, Xu, Fu, and
  Wang]{psvt}
Zhongwei Qiu, Qiansheng Yang, Jian Wang, Haocheng Feng, Junyu Han, Errui Ding,
  Chang Xu, Dongmei Fu, and Jingdong Wang.
\newblock Psvt: End-to-end multi-person 3d pose and shape estimation with
  progressive video transformers.
\newblock In \emph{Proceedings of the IEEE/CVF Conference on Computer Vision
  and Pattern Recognition}, pages 21254--21263, 2023.

\bibitem[Russo(2020)]{regressor}
Alessandro Russo.
\newblock Domain analysis of end-to-end recovery of human shape and pose.
\newblock https://github.com/russoale/hmr2.0, 2020.

\bibitem[Sarkar et~al.(2023)Sarkar, Dave, Medioni, and Biggs]{soy}
Rohan Sarkar, Achal Dave, Gerard Medioni, and Benjamin Biggs.
\newblock Shape of you: Precise 3d shape estimations for diverse body types.
\newblock In \emph{Proceedings of the IEEE/CVF Conference on Computer Vision
  and Pattern Recognition}, pages 3520--3524, 2023.

\bibitem[Sengupta et~al.(2020)Sengupta, Budvytis, and Cipolla]{ssp3d}
Akash Sengupta, Ignas Budvytis, and Roberto Cipolla.
\newblock Synthetic training for accurate 3d human pose and shape estimation in
  the wild.
\newblock In \emph{British Machine Vision Conference (BMVC)}, 2020.

\bibitem[Sengupta et~al.(2021)Sengupta, Budvytis, and Cipolla]{sengupta}
Akash Sengupta, Ignas Budvytis, and Roberto Cipolla.
\newblock Probabilistic estimation of 3d human shape and pose with a semantic
  local parametric model.
\newblock \emph{arXiv preprint arXiv:2111.15404}, 2021.

\bibitem[Shetty et~al.(2023)Shetty, Birkhold, Jaganathan, Strobel, Kowarschik,
  Maier, and Egger]{pliks}
Karthik Shetty, Annette Birkhold, Srikrishna Jaganathan, Norbert Strobel,
  Markus Kowarschik, Andreas Maier, and Bernhard Egger.
\newblock Pliks: A pseudo-linear inverse kinematic solver for 3d human body
  estimation.
\newblock In \emph{Proceedings of the IEEE/CVF Conference on Computer Vision
  and Pattern Recognition}, pages 574--584, 2023.

\bibitem[Sun et~al.(2021)Sun, Bao, Liu, Fu, Black, and Mei]{romp}
Yu Sun, Qian Bao, Wu Liu, Yili Fu, Michael~J Black, and Tao Mei.
\newblock Monocular, one-stage, regression of multiple 3d people.
\newblock In \emph{Proceedings of the IEEE/CVF international conference on
  computer vision}, pages 11179--11188, 2021.

\bibitem[Xu et~al.(2022)Xu, Zhang, Zhang, and Tao]{vitpose}
Yufei Xu, Jing Zhang, Qiming Zhang, and Dacheng Tao.
\newblock Vitpose: Simple vision transformer baselines for human pose
  estimation.
\newblock \emph{Advances in Neural Information Processing Systems},
  35:\penalty0 38571--38584, 2022.

\bibitem[Zhang et~al.(2021)Zhang, Tian, Zhou, Ouyang, Liu, Wang, and
  Sun]{pymaf}
Hongwen Zhang, Yating Tian, Xinchi Zhou, Wanli Ouyang, Yebin Liu, Limin Wang,
  and Zhenan Sun.
\newblock Pymaf: 3d human pose and shape regression with pyramidal mesh
  alignment feedback loop.
\newblock In \emph{Proceedings of the IEEE/CVF international conference on
  computer vision}, pages 11446--11456, 2021.

\end{thebibliography}
